\documentclass{article} 
\usepackage{arxiv}
\setcitestyle{authoryear,open={(},close={)}}
\usepackage{microtype}
\usepackage{graphicx}
\usepackage{subfigure}
\usepackage{booktabs}
\usepackage{fullpage}

\usepackage{amsmath,amsfonts,bm}

\def\eqref#1{equation~\ref{#1}}

\def\1{\bm{1}}

\DeclareMathAlphabet{\mathsfit}{\encodingdefault}{\sfdefault}{m}{sl}
\SetMathAlphabet{\mathsfit}{bold}{\encodingdefault}{\sfdefault}{bx}{n}

\usepackage{url}
\usepackage{graphicx}
\usepackage{wrapfig}
\usepackage{algorithm}
\usepackage[noend]{algorithmic}
\usepackage{bbm}
\usepackage{amsmath}
\usepackage{amssymb}
\usepackage{mathtools}
\usepackage{amsthm}
\usepackage{enumitem}
\usepackage{natbib}
\usepackage{xcolor}
\usepackage{multirow}
\usepackage{subcaption}
\usepackage{setspace}
\usepackage{enumitem}
\usepackage{color, colortbl}
\usepackage{mathtools}
\usepackage{cleveref}

\theoremstyle{plain}

\theoremstyle{definition}

\theoremstyle{remark}

\usepackage{xcolor}         
\usepackage{multirow}
\usepackage{multicol}
\usepackage{graphicx}
\usepackage{amsmath}
\usepackage{amsthm}
\usepackage{colortbl}  
\usepackage{xcolor}
\usepackage{bm}
\usepackage{enumitem}
\usepackage{wrapfig}
\usepackage{amssymb}
\usepackage{calc}
\usepackage{algorithm}
\usepackage{algorithmic}
\title{\textbf{Pre-Trained Video Generative Models \\ as World Simulators}}

\author{
{\large Haoran He$^{1}$ ~ Yang Zhang$^{2}$ ~ Liang Lin$^{3}$ ~ Zhongwen Xu$^{4}$ ~ Ling Pan$^{1}$\thanks{Correspondence to: Ling Pan \href{lingpan@ust.hk}{(lingpan@ust.hk)}.}} \\
 \\
{\large $^{1}$ Hong Kong University of Science and Technology $^{2}$ Tsinghua University}\\
{\large $^{3}$ Sun Yat-sen University $^{4}$ Tencent AI Lab} \\  
}

\begin{document}

\maketitle

\begin{abstract}
Video generative models pre-trained on large-scale internet datasets have achieved remarkable success, excelling at producing realistic synthetic videos. However, they often generate clips based on static prompts (e.g., text or images), limiting their ability to model interactive and dynamic scenarios. In this paper, we propose \textbf{D}ynamic \textbf{W}orld \textbf{S}imulation (DWS), a novel approach to transform pre-trained video generative models into controllable world simulators capable of executing specified action trajectories. To achieve precise alignment between conditioned actions and generated visual changes, we introduce a lightweight, universal action-conditioned module that seamlessly integrates into any existing model. Instead of focusing on complex visual details, we demonstrate that consistent dynamic transition modeling is the key to building powerful world simulators. Building upon this insight, we further introduce a motion-reinforced loss that enhances action controllability by compelling the model to capture dynamic changes more effectively. Experiments demonstrate that DWS can be versatilely applied to both diffusion and autoregressive transformer models, achieving significant improvements in generating action-controllable, dynamically consistent videos across games and robotics domains. Moreover, to facilitate the applications of the learned world simulator in downstream tasks such as model-based reinforcement learning, we propose prioritized imagination to improve sample efficiency, demonstrating competitive performance compared with state-of-the-art methods.
\end{abstract}
\section{Introduction}

The field of video generation has experienced remarkable progress in recent years, with models such as \citet{videoworldsimulators2024,opensora,polyak2024movie,yang2024cogvideox,veo2024} demonstrating an exceptional ability to generate high-fidelity and temporally consistent videos conditioned on various inputs, most notably text and initial frames. However, these models are limited to support interactive simulation scenarios, as they are trained for one-shot generation with static prompts, lacking frame-level interactivity and frame-to-frame dynamic modeling. To fill this gap, the community is increasingly focusing on building action-conditioned video models~\cite{yang2023learning,bruce2024genie,xiang2024pandora,wu2024ivideogpt,valevski2024diffusion,oasis2024,2025gamegen,yang2024playable}.

These action-conditioned models effectively act as interactive environment simulators (``world models'' or ``world simulators''), which leverage advanced transformers or diffusion model architectures to predict future visual outcomes based on the agent's actions. Their goal is to encapsulate an understanding of the underlying dynamic transitions and commonsense knowledge about how the world works, enabling action-driven imagination analogous to the human cognition process. These world models open exciting possibilities, particularly in model-based reinforcement learning (MBRL), where agents can learn new skills more efficiently by interacting with world models, avoiding the risks and costs that arise from real-world trials.

In this work, we review recent advances in interactive world simulators, highlighting key challenges that currently limit their broader adoption. (\romannumeral1) These models often require vast computational resources for training from scratch. For example, Genie \cite{bruce2024genie} required 125k training steps on 256 TPUv5p cores (roughly equivalent to 226 NVIDIA A100 GPUs) to learn a relatively simple Platformer game simulator. Similarly, GameNGen~\cite{valevski2024diffusion} consumed 700k steps with 128 TPU-v5e cores. (\romannumeral2) While fine-tuning a pre-trained video generative model offers a more efficient alternative, existing approaches~\citep{rigter2024avid,yu2025gamefactory} are inherently architecture-dependent. This poses challenges to adapting these methods across different model architectures to benefit from rapid advances in video generation architectures, as each new model architecture requires substantial engineering efforts for adaptation. (\romannumeral3) Unlike general video generation tasks, action-conditioned world simulators require precise capture of fine-grained dynamic changes~\citep{FangqiIRASim2024,yang2023learning}, which requires frame-level action alignment. This requirement is crucial for applications in model-based reinforcement learning, where capturing frame-to-frame dynamic/motion changes takes precedence over modeling static visual elements (e.g., background, object details).

\begin{wrapfigure}{r}{.5\linewidth}
    \centering
    \includegraphics[width=1.\linewidth]{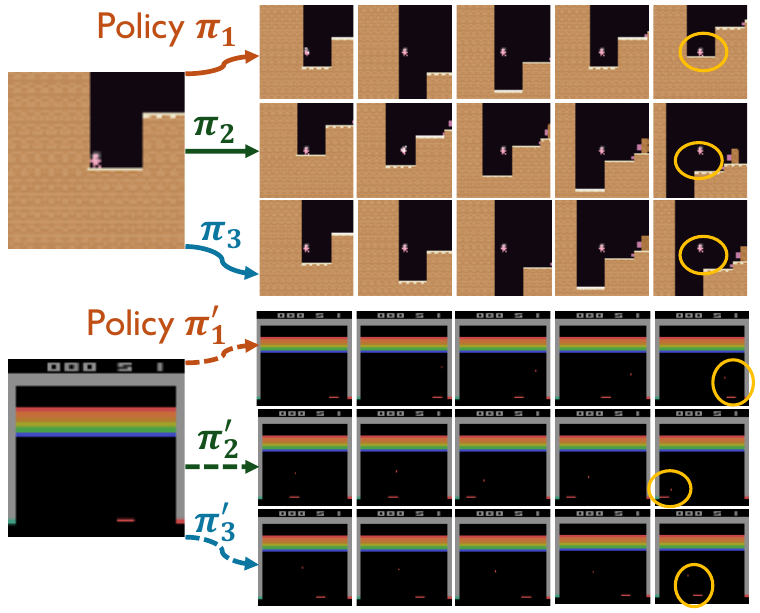}
    \vspace{-2.em}
    \caption{Our fine-tuned video generative model can serve as an effective world simulator for interacting with different policies and generating diverse trajectories. Different policies can lead to different terminal states, which are highlighted with yellow circles.}
    \vspace{-1.em}
    \label{fig:demo_planning}
\end{wrapfigure}
To address the aforementioned challenges, we propose a novel framework, \textbf{D}ynamic \textbf{W}orld \textbf{S}imulation (DWS), which is a unified, architecture-agnostic approach for efficiently converting pre-trained video generative models into world simulators. By leveraging pre-trained priors learned from internet-scale datasets, the fine-tuned world simulators can demonstrate a basic understanding of physical rules and commonsense knowledge. However, they are not inherently equipped for interactive simulation and lack the key mechanisms for precise frame-level action conditioning. 
DWS introduces a minimalist yet powerful add-on action-conditioned module that improves frame-level action awareness while maintaining architectural flexibility.
This module comprises just two linear layers and can be integrated into any network architecture through carefully designed scale and shift operations. It also strengthens the alignment between predicted visual changes and conditioned actions. Furthermore, we observe that traditional supervised learning loss functions lead video models to focus uniformly across all visual elements and complex details, including static backgrounds and irrelevant details, compromising their ability to capture the frame-to-frame dynamic/motion changes crucial for world simulator construction.
DWS presents a motion-reinforced loss to address the dynamic modeling challenge, a simple yet effective method to explicitly prioritize the modeling of inter-frame changes during training, resulting in significantly improved temporal consistency and more reliable dynamic predictions. As shown in Fig.~\ref{fig:demo_planning}, the learned world simulators by DWS can interact with diverse policies across different domains while maintaining accurate dynamic prediction and action responsiveness. Finally, to enhance the practical utility of world simulators in model-based reinforcement learning, we introduce prioritized imagination, a novel sampling strategy that focuses on the most informative transitions rather than wasting computational resources on well-understood state transitions, leading to improved sample efficiency during agent-world model interactions.

We summarize the contributions of this paper as follows: (\romannumeral1) We introduce DWS, a novel and architecture-agnostic framework that effectively converts pre-trained video generative models to world simulators with low training costs, leveraging the pre-trained prior knowledge for physics grounding. (\romannumeral2) We introduce two simple yet effective techniques: 
a lightweight action-conditioned module that enables precise frame-level control and improve action-following ability, and a motion-reinforced training that redirects model attention from static visual details to action-induced dynamic changes for improving temporal, dynamic consistency. 
(\romannumeral3) We advance the practical utility of world simulators in model-based reinforcement learning through prioritized imagination, which improves sample efficiency and policy performance when applying the trained world simulators to downstream model-based RL. (\romannumeral4) Through comprehensive evaluation across challenging game and robotics tasks, we demonstrate that DWS significantly improves the quality and dynamic consistency of generated action-conditioned videos. Our DWS-trained world simulators with prioritized imagination also enable more efficient and effective learning compared to previous state-of-the-art MBRL approaches~\citep{hafner2023dreamerv3,alonso2024diamond}.
\vspace{-0.5em}
\section{Related Work}
\textbf{Video World Models.} With the development of internet-scale datasets~\citep{Bain21,chen2024panda70m} and advanced model architecture~\citep{peebles2023dit,videoworldsimulators2024}, significant progress has been made in realistic video generation conditioned on text descriptions and
initial frames~\citep{blattmann2023SDVideo,lin2024opensoraplan, ma2024latte, yang2024cogvideox,opensora}. Building upon these foundations, current research has increasingly focused on action-controllable video generation, aiming to develop generalist world simulators~\citep{xiang2024pandora,yang2023learning,bruce2024genie,feng2024matrix,valevski2024diffusion,genie2,FangqiIRASim2024,2025gamegen} that can effectively model both physical dynamics and action consequences. However, these models typically require training from scratch on large-scale datasets and involve millions (or billions) of parameters, resulting in substantial computational overhead and slow inference speed. In contrast, we propose to adapt publicly available pre-trained video generative models~\citep{opensora,wu2024ivideogpt} into action-driven world simulators. Our proposed fine-tuning approach, DWS, achieves efficient model adaptation while requiring minimal computational overhead, significantly reducing both training costs and inference latency. Concurrent works~\citep{rigter2024avid,yu2025gamefactory} have also explored leveraging pre-trained models for action-conditioned video generation. However, their investigations are limited to diffusion-based models, and neither work validates the effectiveness of their approaches in facilitating downstream tasks such as model-based RL.

\noindent \textbf{Model-Based RL}
Model-based RL aims to build world models in which the trial-and-errors can take place without real cost. With a sufficiently accurate world model, agents can develop imagination abilities, allowing them to simulate interactions and generate synthetic experience data. This simulated data can then be leveraged to learn optimal policies for diverse decision-making tasks, effectively reducing the need for real-world interactions. \citet{sutton1991dyna} introduce the first general framework for model-based RL, highlighting the utility of an estimated dynamics model in facilitating the training of value functions and policies~\citep{Sutton1998ReinforcementL}. Recent years have witnessed remarkable progress in model-based methods for learning complex environmental dynamics, such as video games and visual control tasks, consistently outperforming their model-free counterparts. For example, built upon Recurrent State Space Models (RSSM)~\citep{hafner2019PlaNet}, which explicitly decouple the deterministic and stochastic components of environmental dynamics, the Dreamer series has demonstrated impressive performance across diverse domains, including Atari games~\citep{atari}, DeepMind Control Suite~\citep{tassa2018deepmind}, and Minecraft. 
To address the limitation of RNNs in expressing complex patterns, recent works have explored leveraging transformer models for enhanced sequence modeling and long-term dependency capture~\citep{micheli2023iris,robine2023twm,zhang2023storm,zhang2024marie}, and incorporating diffusion models to better represent multi-modal distributions in dynamic learning~\citep{ding2024dwm,alonso2024diamond}. However, although these works also employ transformer or diffusion models for world model learning, they predominantly rely on training from scratch and fail to leverage pre-trained knowledge for enhanced dynamics understanding, making them overly task-specific and limiting their ability to generalize across diverse tasks. Furthermore, while existing methods treat all imagined samples with uniform importance during training, our proposed DWS introduces a novel prioritization mechanism that selectively focuses on significant samples, thereby improving sample efficiency. 
\vspace{-0.5em}
\section{Preliminaries}
\subsection{Problem Formulation}
\vspace{-0.5em}
The conditional video generation framework
can be adaptable to instantiate a world simulator (or a world model)~\citep{yang2023learning}. The world model takes in some action as input and produces the visual consequence of the action as output, which aims to simulate the environment. This environment can be represented as a Partially Observable Markov Decision Process (POMDP), encapsulated within the tuple $(\mathcal{S},\mathcal{O},\mathcal{\phi},\mathcal{A},p,r,\gamma)$. Here, $\mathcal{S}$ is the state space, and $\mathcal{O}$ is the observation space which only provides incomplete information of $\mathcal{S}$. At each timestep $t$, the agent chooses an action $a_t$ by following a policy $\pi:\mathcal{O}\to\Delta_\mathcal{A}$, the environment updates the state following the dynamics, $s_{t+1}\sim p(s_{t+1}|s_t,a+t)$, the next observation $o_{t+1}=\phi(s_{t+1})$ is received and a scalar reward $r_t$ is computed as $R(s_t,a_t,s_{t+1})$. The goal of the agent is to learn a policy $\pi^*=\arg\max_\pi\mathbb{E}_{a_t \sim \pi}\big[\sum\nolimits_{t=0}^{\infty}\gamma^t r_t\big]$ by maximizing the $\gamma$-discounted cumulative rewards. 

A well-trained world model can replace the environment to interact with the agent, and thus benefit downstream policy learning by providing infinite experiences. Concretely, given a history observation $o_{T_0}$, at each timestep $t=T_0,\cdots,T-1$, the agent takes an action $a_t$ based on its policy and previous imagined observations, and then the world model predicts the transition $p(o_{t+1}, r_{t+1} | o_t, a_t)$ to feedback the agent. 
\subsection{Pre-Trained Video Generative Models}
By formulating learning world models for visual control as an interactive video generation problem, we can harness the widely available video data, which embeds broad knowledge that is generalizable across different domains~\citep{pmlr-v235-yang24z}. Video data not only contains semantic visual details but also includes motion movements that capture the dynamic rules in the physical world. However, training such video world models on internet-scale video datasets from scratch is expensive and time-consuming. We propose to fine-tune pre-trained advanced video models to enable them to simulate interactions. Specifically, we adopt two different pre-trained video generative models as our base models, which are diffusion models, i.e., Open-Sora~\citep{opensora}, and autoregressive transformer models, i.e., iVideoGPT~\citep{wu2024ivideogpt}. Open-Sora is a kind of rectified flow-based diffusion model that is fully open-sourced and pre-trained on millions of internet videos. We consider using it because it requires only a few sampling steps, benefiting from flow-matching training. A recent work named iVideoGPT is an autoregressive transformer built upon LLaMA~\citep{touvron2023llama} architecture. It compresses training data from different modalities (e.g., including visual outcomes, actions, and rewards) into a sequence of tokens for interactive video prediction.

\vspace{-0.5em}
\section{Method}
In this section, we first introduce our proposed action-conditioned module in §\ref{sec:module}, and the motion-reinforced loss for enhancing dynamic modeling in §\ref{sec:motion}. After presenting methods for fine-tuning general video generative models into world simulators, we introduce the prioritized imagination technique for improving model-based reinforcement learning performance in §\ref{sec:mbrl}.

\subsection{Action-Conditioned Module}
\label{sec:module}
Recent video generative models have achieved significant success in generating realistic videos correlated with conditioned text prompts or initial frames. These text-to-video tasks operate with static prompts that globally describe the entire video without specifying what the next frames should be. This design paradigm, while effective for general video generation, presents fundamental challenges when adapting them as world models that aim to simulate action-rich interactions, where the conditions are frame-level, fine-grained action trajectories.
To address this requirement and ensure each generated frame matches its corresponding action in the trajectory, we leverage an action-conditioned module that conditions the generation of each frame by its corresponding action individually. Unlike previous text-to-video models that compress the entire action trajectory into a single embedding, our approach, similar to IRASim~\citep{FangqiIRASim2024}, implements a more granular action encoding mechanism. We introduce a lightweight add-on module, consisting of two linear layers within each transformer block, to encode individual actions separately. This design ensures that each frame's content is directly modulated by its corresponding action, rather than being guided by a global description, and leads to a direct correspondence between actions and generated frames.

\textbf{Action Representation.} 
A key challenge in adapting video generative models for action-based control lies in the representation of actions.
In discretized action spaces, actions are typically represented as integer values, which lack the rich semantic context present in text prompts used in traditional text-to-video models. This semantic gap can limit the model's ability to interpret and respond to different actions effectively. To bridge this gap, we propose to represent the actions using language templates that depict the meanings of the actions. Specifically, given an action trajectory $y=\{a_t,a_{t+1},\cdots,a_{t+H-1}\}$, where $H$ is the horizon of the trajectory, we develop a mapping function $\psi$ to translate abstract action integers into meaningful languages, i.e., $\psi:\mathcal{A}\to L$, where $L$ is the language space. This mapping enables us to leverage the text encoder in pre-trained video generative models to obtain rich feature embeddings $c\in \mathbb{R}^{n_H\times n_d}$, where $n_H$ and $n_d$ represent the horizon and the dimension of each token respectively. For continuous action spaces, following \citet{wu2024ivideogpt,FangqiIRASim2024}, we use a trainable linear action embedder to directly generate feature embeddings $c$ without language translation.

\textbf{Frame-Level Condition.} In the context of video generative models serving as world simulators, precise temporal control is important as each action should directly modulate the visual content of its subsequent frame. To explicitly model and enhance this action-frame correspondence, we incorporate a frame-level action-conditioning module within each transformer block, drawing inspiration from IRASim~\citep{FangqiIRASim2024}. While IRASim's implementation was limited to specific diffusion models with temporal-spatial transformer architectures, we significantly extend this concept by developing a versatile add-on module that generalizes across different model architectures, including both diffusion-based and transformer-based frameworks. Therefore, our design offers enhanced architectural flexibility and broader applicability. Our minimalist architecture, implemented with just two linear layers, enables lightweight integration and efficient fine-tuning with minimal computational overhead.

\begin{wrapfigure}{r}{.55\linewidth} \vspace{-.1in}
     \centering
     
     \includegraphics[width=1.\linewidth]{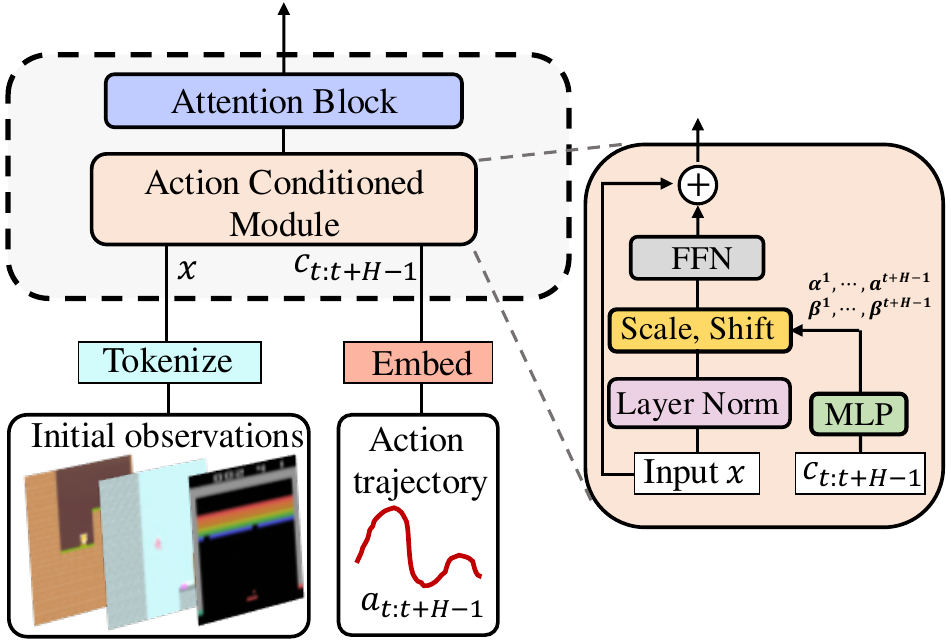}
     \caption{Illustration of the action-conditioned module, which can be incorporated with any type of transformer attention block.}
     \vspace{-1.2em}
     \label{fig:architecture}
 \end{wrapfigure}

Specifically, for each video embedding $x\in\mathbb{R}^{T\times C\times H\times W}$, we process them as follows before feeding them into the transformer block:
\begin{equation}
    x^i = x^i + {\rm FFN}({\rm LayerNorm}(x^i)\times (1+\alpha^i) + \beta^i),
\end{equation}
where $\alpha^i$ and $\beta^i$ denote the scale and shift parameters for the $i$-th frame. They are regressed from the action embedding $c^i$. We illustrate our proposed module in Fig.~\ref{fig:architecture}, which can be seamlessly integrated into any network block (e.g., attention block).
 
\subsection{Motion-Reinforced Loss}
\label{sec:motion}
Traditional video generative models commonly employ the squared $\mathit{l}_2$ distance (for rectified flow~\citep{liu2022flow,opensora} in the continuous space) or cross-entropy loss (for next-token-prediction transformers~\citep{vaswani2017attention,wu2024ivideogpt} in discrete space) as their training objectives. While these training objectives have demonstrated effectiveness in general video generation tasks~\citep{opensora,lin2024opensoraplan,videoworldsimulators2024,yan2021videogpt,tian2024visual}, they typically consider each pixel equally, which may compromise the model's ability to capture action-dependent state changes. Therefore, it is inefficient for them to function effectively as world simulators for RL agents, where accurate modeling of dynamic transitions is more crucial for learning than maintaining high-fidelity background details. This limitation arises because RL agents predominantly learn from action-induced state changes rather than static visual elements.

To tackle this problem and enable more precise dynamic modeling, we introduce a new motion-reinforced loss
to improve the action-following ability of video models. 
At each training step, we sample a random batch of ground-truth video embeddings $x=\{x^0,\cdots,x^k,\cdots,x^j,x^{H-1}\}$, where $H$ denotes the horizon of the video clips. We then compute the differences $\omega=\cup_{i=0}^{H-1}\omega_i$ between consecutive frames, denoted as 
\begin{equation}
    \omega_{i+1} = c^{{\rm Softmax}(|x_{i+1}-x_{i}|)},
\end{equation}
where $\omega_i\to[1,c]$, and we set $\omega_0=1$ for the initial frame $x^0$ since it serves as a conditioned frame. Here, $c$ denotes a hyperparameter that modulates the motion-reinforced strength. After obtaining $\omega$, we integrate it as pixel-wise weights into the supervised training loss. The resulting motion-reinforced loss function can be formulated as:
\begin{equation}
    \mathcal{L}_{\rm motion} =  \mathcal{L}_{\rm prev}*\omega,
\end{equation}
where $\mathcal{L}_{\rm prev}$ represents either the original MSE loss used in diffusion models, or the cross-entropy loss function in transformer-based architectures. We include more implementation details in Appendix~\ref{appendix:motion_loss}.
Through this formulation, pixels that change across frames will have a greater impact on loss backpropagation. These pixels typically correspond to motion-related elements in videos, which undergo continuous changes, while the background, which remains stable, will have less influence during training. This mechanism inherently attenuates the impact of static background elements that contribute minimally to action-conditioned prediction.

By emphasizing dynamic transitions, our approach enhances the world model's capability to capture action-state causal relationships, thereby facilitating more effective policy learning in reinforcement learning contexts.

\begin{figure}
    \centering
    \subfigure[frame-to-frame differences]{\includegraphics[width=0.42\linewidth]{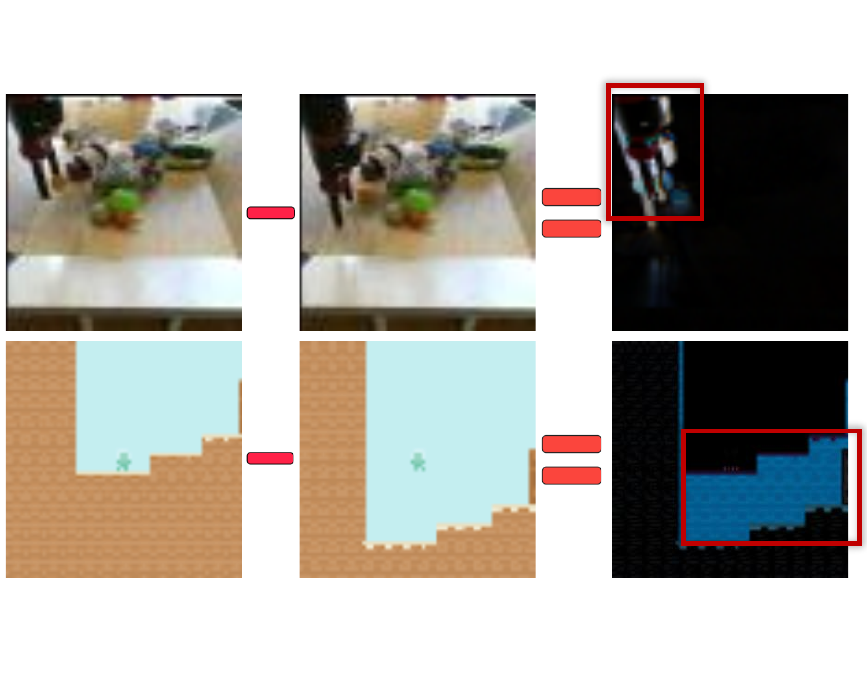}}
    \subfigure[SFT loss $\mathcal{L}_{\rm prev}$ comparison]{\includegraphics[width=0.42\linewidth]{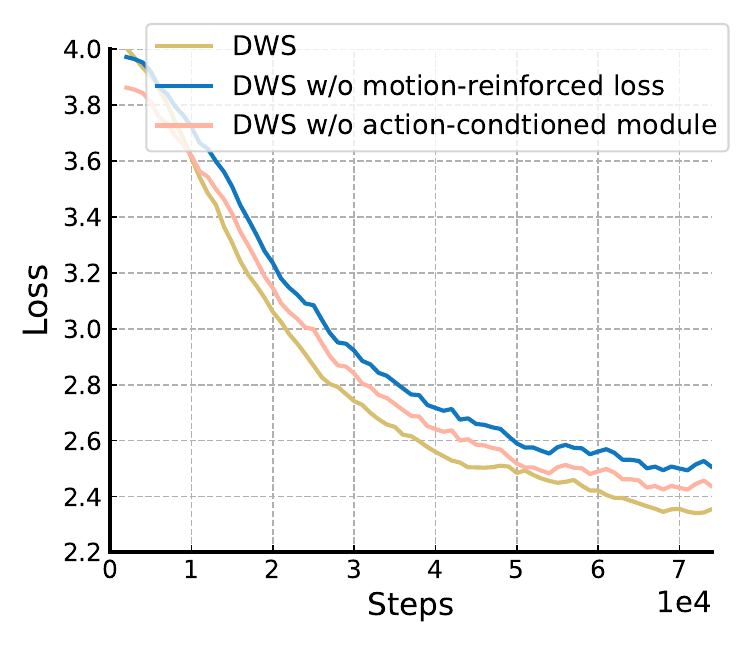}}
    \caption{Motion-reinforced loss enhances the action-controllability of video generative models, helping capture dynamically changing contents. }
    \label{fig:motion_loss}
    \vspace{-1.5em}
\end{figure}

As illustrated in Figure~\ref{fig:motion_loss}(a), inter-frame differences predominantly correspond to motion-related and dynamic elements in videos, leading $\omega$ to assign higher weights to these pixels during the training process. Figure~\ref{fig:motion_loss}(b) presents the SFT loss ($\mathcal{L}_{\rm prev}$) curves from fine-tuning iVideoGPT~\citep{wu2024ivideogpt} on the BAIR dataset~\cite{ebert2017self}, comparing different variants of our proposed method. The empirical results demonstrate that both the motion-reinforced loss and the action-conditioned module are crucial components, as the absence of either component significantly degrades the model's performance in predicting action-conditioned videos with frame-to-frame dynamics.

Therefore, the video generative models fine-tuned by $\mathcal{L}_{\rm motion}$ will focus more on the dynamic/motion prediction instead of complex visual details that are challenging to learn. Moreover, dynamic/motion consistency is more important than static background details for world simulators, since simulators are required to predict visual outcomes conditioned on actions.

\vspace{-0.5em}
\subsection{Model-Based Reinforcement Learning}
\label{sec:mbrl}
Given the video-based world simulators fine-tuned from pre-trained video generative models, one of the most promising applications is to utilize them as world models for policy learning in model-based reinforcement learning (MBRL). In MBRL, the agent optimizes a policy to maximize the cumulative rewards by interacting with the trained world models which improves sample efficiency. 

To enable effective policy training in MBRL, the world model should predict both transition dynamics and rewards. We now complete our world model with a reward prediction model. Since estimating the reward is a scalar prediction problem, we introduce a separate model $R_\psi$ consisting of linear layers, self-attention blocks, and cross-attention blocks to estimate the reward given past observations and actions. The RL agent involves an actor-critic network parameterized by
a shared CNN backbone that branches into separate policy and value heads. Building upon the MBPO framework~\citep{janner2019trust,wu2024ivideogpt}, we augment
the replay buffer with synthetic rollouts to train a standard actor-critic RL algorithm. We adopt PPO~\citep{schulman2017proximal} as our base algorithm and follow \citet{shengyi2022the37implementation} for implementation. We include more implementation details in Appendix~\ref{appendix:mbrl}.

\textbf{Prioritized Imagination.} Imagination by a world model needs to start from initial observations, which are sampled from the experiences collected from the environment. 
These initial states serve as starting points for the world model to generate synthetic trajectories.
Previous MBRL methods~\citep{Hafner2020Dream,hafner2020dreamerv2,hafner2023dreamerv3,alonso2024diamond,micheli2023transformers} employ uniform sampling of initial observations for imagination. However, this strategy neglects the varying importance of different states for policy learning, leading to learning inefficiency.
We highlight that imagined transitions originating from different initial observations exhibit substantial heterogeneity in their importance and task relevance for MBRL policy optimization.
To better unlock the world simulation ability of fine-tuned video generative models, we propose a prioritized imagination method that selectively focuses on more valuable transitions. Our key insight is that initial observations leading to transitions with higher learning potential and learn-ability should be sampled more frequently. We maintain a buffer $\mathcal{B}$ to store observations encountered during the interaction with environments, and prioritize initial observations with high expected learning progress, which is measured by the magnitude of their TD loss. 
This prioritization mechanism ensures more efficient utilization of the world model by concentrating imagination resources on states that yield more substantial contributions to policy learning.
The overall pseudo-code is presented in Algorithm~\ref{alg:mbpo}.

\section{Experiments}
\begin{figure*}[htbp]
\begin{minipage}{1.\textwidth}
    \centering
    \includegraphics[width=1.\linewidth]{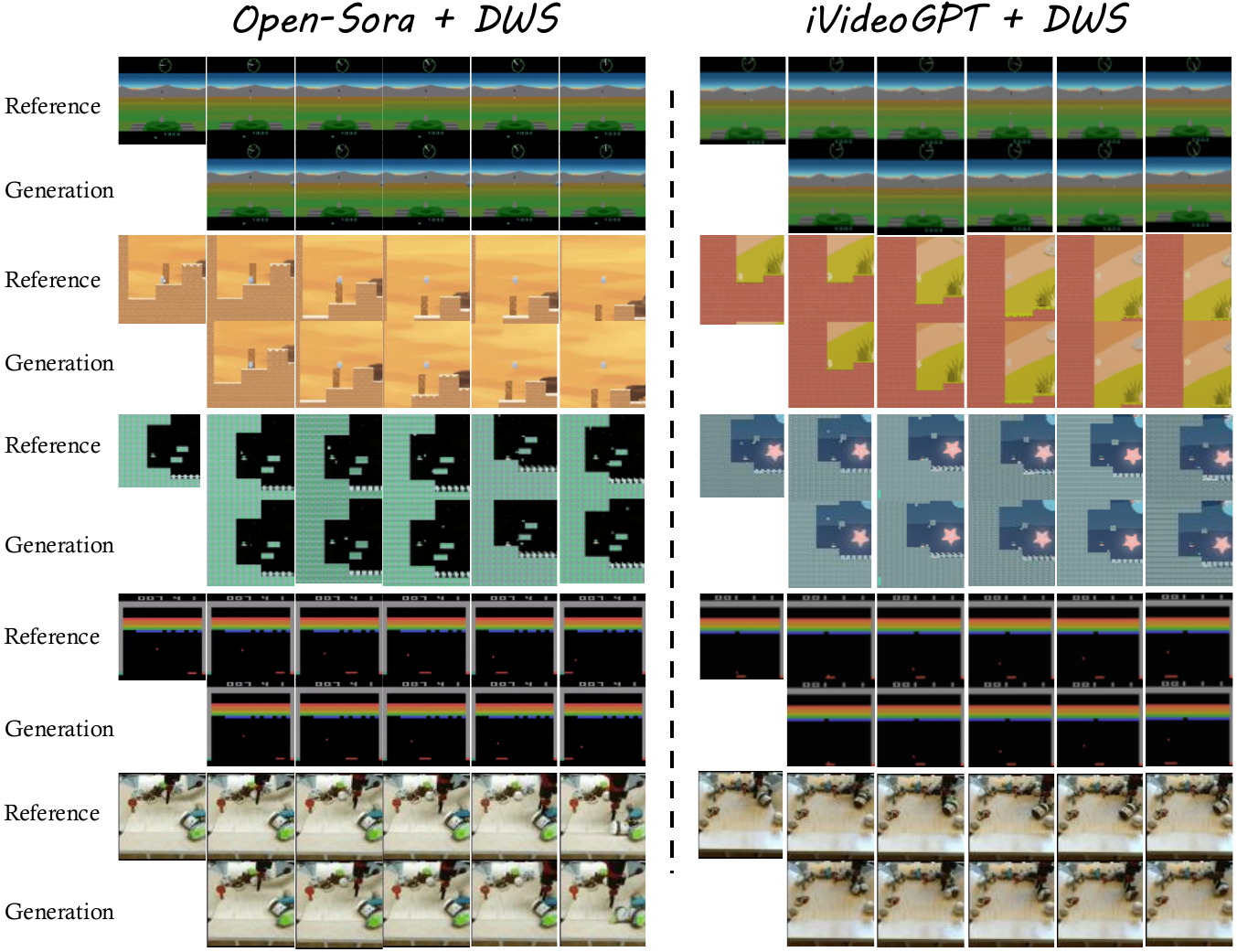}
    \vspace{-1.3em}
    \caption{The qualitative results of generated videos for different domains, including games and robotics environments. Given initial observations and conditioned actions, we observe that Open-Sora and iVideoGPT fine-tuned by our proposed method significantly improve dynamic world modeling ability.}
    \label{fig:demos}
\end{minipage}
\end{figure*}
In this section, we evaluate the video world model fine-tuned by our proposed method from two critical perspectives: (1) action simulation capability, assessed through the quality of action-conditioned video prediction, and (2) the effectiveness in both online and offline model-based reinforcement learning, quantified by the cumulative return across tasks.
\subsection{Action-Conditioned Simulation} 
To evaluate the effectiveness of DWS in enhancing action-conditioned video prediction for world simulation, we conduct experiments using two architecturally distinct pre-trained video generative models: Open-Sora~\citep{opensora}, which employs a diffusion-based architecture, and iVideoGPT~\citep{wu2024ivideogpt}, which utilizes an autoregressive architecture. 
For computationally efficiency, we utilize a compressed version of Open-Sora with 12 layers (approximately 280M parameters) instead of the original 1.1B model. The details and our setup of these base models can be found in Appendix~\ref{appendix:base_models}. The evaluation is performed across three diverse datasets: the BAIR dataset~\citep{ebert2017self} featuring continuous action spaces, and both Procgen~\citep{cobbe2020leveraging} and Atari~\citep{bellemare2013arcade,atari} datasets incorporating discrete action spaces.

\textbf{Experiment Setup.} The BAIR robot pushing dataset which is about a robotic arm manipulating various objects consists of 43k training and 256 test videos, 
Following previous works~\citep{yan2021videogpt,gupta2022maskvit}, we predict 15 frames from a single initial frame. In this benchmark, we compare against a variety of video prediction models, including diffusion~\citep{opensora,voleti2022mcvd}, masked~\citep{yu2023magvit,gupta2022maskvit}, and autoregressive models~\citep{wu2024ivideogpt,yan2021videogpt}. For the Procgen dataset, we evaluate DWS on two platformer games, i.e., namely Coinrun and Ninja. For the Atari dataset, we assess performance on two Atari games: Breakout and Battle Zone. We collect 1M transition steps for each game, encompassing RGB observations, actions, and rewards. The resolution is $64\times 64$ for both BAIR and Procgen videos, while the resolution for Atari videos is $84\times84$. More details regarding dataset collection are provided in Appendix~\ref{appendix:dataset}. 

\textbf{Metrics.} We adopt four widely-used metrics to measure the quality of predicted videos: across four metrics: 1): FVD~\citep{unterthiner2018fvd} quantifies the statistical similarity between reference and generated video distributions by computing the Fréchet distance between feature representations extracted from a pre-trained network~\citep{carreira2017i3d}; 2) PSNR~\citep{HuynhThu2008psnr} measures the pixel-wise fidelity between reference and generated frames, which is computed as the logarithmic ratio between the maximum possible pixel value and the mean squared error; 3) SSIM~\citep{HuynhThu2008psnr} evaluates the structural information preservation in generated frames, which incorporates luminance, contrast, and structural comparisons; 4) LPIPS~\citep{zhang2018lpips} leverages a pre-trained deep neural network~\citep{simonyan2014vgg} to assess image similarity by computing distances in deep feature space.

\begin{table}[htbp]
\vspace{-1.4em}
\begin{minipage}[c]{0.48\textwidth}

    \caption{Video prediction results on the BAIR robot pushing dataset. LPIPS and SSIM scores are scaled by 100 for convenient display. 
    }
    \vspace{5pt}
    \label{tab:video_pred}
    
    \centering
    \small
    \setlength{\tabcolsep}{2.3pt}
   \scalebox{0.9}{
    \begin{tabular}{lllll}
    \toprule
    \textbf{BAIR} \cite{ebert2017self} & FVD$\downarrow$ & PSNR$\uparrow$ & SSIM$\uparrow$ & {\hspace{-3pt}LPIPS$\downarrow$ \hspace{-5pt}} \\ \midrule
      VideoGPT \cite{yan2021videogpt}  &  103.3   &  -   &   -   &   -   \\
      MaskViT  \cite{gupta2022maskvit} &  93.7   &   -   &   -   &   -   \\
      FitVid \cite{babaeizadeh2021fitvid}  &  93.6   &   -  &  -    &  -    \\
       MaskViT \cite{gupta2022maskvit}  & 70.5    &  -    &  - &  -     \\ 
      MCVD \cite{voleti2022mcvd}   &  89.5   &  16.9    &  78.0    &    -   \\
    
      MAGVIT \cite{yu2023magvit}  &   62.0  &  19.3   &   78.7   &  12.3     \\ 
      \midrule
      Open-Sora~\citep{opensora}&92.1&21.5&84.7&8.6\\
      \rowcolor[rgb]{0.9,1.0,0.9}Open-Sora\textbf{+DWS}(Ours) & \textbf{81.3}  & \textbf{22.4} & \textbf{87.8} &  \textbf{6.2}\\ 
      \midrule
      iVideoGPT \cite{wu2024ivideogpt}   & 60.8    &  24.5    &  90.2 &  5.0     \\ 
      \rowcolor[rgb]{0.9,1.0,0.9}iVideoGPT\textbf{+DWS} (Ours)&\textbf{59.6}&\textbf{25.8}&\textbf{91.6}&\textbf{4.7}\\
      \bottomrule
    \end{tabular}
    }
    \end{minipage}
    \begin{minipage}[c]{0.48\textwidth}
   
    \vspace{-1.9em}
    \caption{Video prediction results on Atari and Procgen game domains.}
    \vspace{5pt}
    \label{tab:game_video_pred}
    \centering
    \small
    \setlength{\tabcolsep}{2.3pt}
    \scalebox{0.9}{
    \begin{tabular}{l|clllll}
    \toprule
    Domain&Method& FVD$\downarrow$ & PSNR$\uparrow$ & SSIM$\uparrow$ & {\hspace{-3pt}LPIPS$\downarrow$ \hspace{-5pt}} \\ 
      \midrule
      \multirow{4}{*}{\textbf{Atari}}&Open-Sora~\citep{opensora}&27.8&31.6&95.2&6.4\\
      &\cellcolor[rgb]{0.9,1.0,0.9}Open-Sora\textbf{+DWS}(Ours) & \cellcolor[rgb]{0.9,1.0,0.9}\textbf{16.3}  & \cellcolor[rgb]{0.9,1.0,0.9}\textbf{35.8} & \cellcolor[rgb]{0.9,1.0,0.9}\textbf{98.0} &  \cellcolor[rgb]{0.9,1.0,0.9}\textbf{4.9}\\ \cline{2-6}
      
      &iVideoGPT \cite{wu2024ivideogpt}   & 11.5    &  39.1    &  97.3 &  3.1     \\ 
      &\cellcolor[rgb]{0.9,1.0,0.9}iVideoGPT\textbf{+DWS} (Ours) &\cellcolor[rgb]{0.9,1.0,0.9}\textbf{9.1}&\cellcolor[rgb]{0.9,1.0,0.9}\textbf{43.4}&\cellcolor[rgb]{0.9,1.0,0.9}\textbf{98.2}&\cellcolor[rgb]{0.9,1.0,0.9}\textbf{1.2}\\
      \midrule
      \multirow{4}{*}{\textbf{Procgen}}&Open-Sora~\citep{opensora}&37.6&22.7&74.7&13.6\\
      &\cellcolor[rgb]{0.9,1.0,0.9}Open-Sora\textbf{+DWS} (Ours) & \cellcolor[rgb]{0.9,1.0,0.9}\textbf{28.2}  & \cellcolor[rgb]{0.9,1.0,0.9}\textbf{23.9} & \cellcolor[rgb]{0.9,1.0,0.9}\textbf{76.1} &  \cellcolor[rgb]{0.9,1.0,0.9}\textbf{12.8}\\ \cline{2-6}
      
      &iVideoGPT \cite{wu2024ivideogpt}   & 24.0    &  25.3    &  77.6 &  12.1     \\ 
      &\cellcolor[rgb]{0.9,1.0,0.9}iVideoGPT\textbf{+DWS} (Ours) &\cellcolor[rgb]{0.9,1.0,0.9}\textbf{21.9}&\cellcolor[rgb]{0.9,1.0,0.9}\textbf{26.2}&\cellcolor[rgb]{0.9,1.0,0.9}\textbf{80.4}&\cellcolor[rgb]{0.9,1.0,0.9}\textbf{10.3}\\
      \bottomrule
    \end{tabular}
    }
    
   
    \end{minipage}
    \end{table}

\begin{wrapfigure}{r}{0.55\textwidth}
    \centering
    \vspace{-1.5em}
    \includegraphics[width=1.\linewidth]{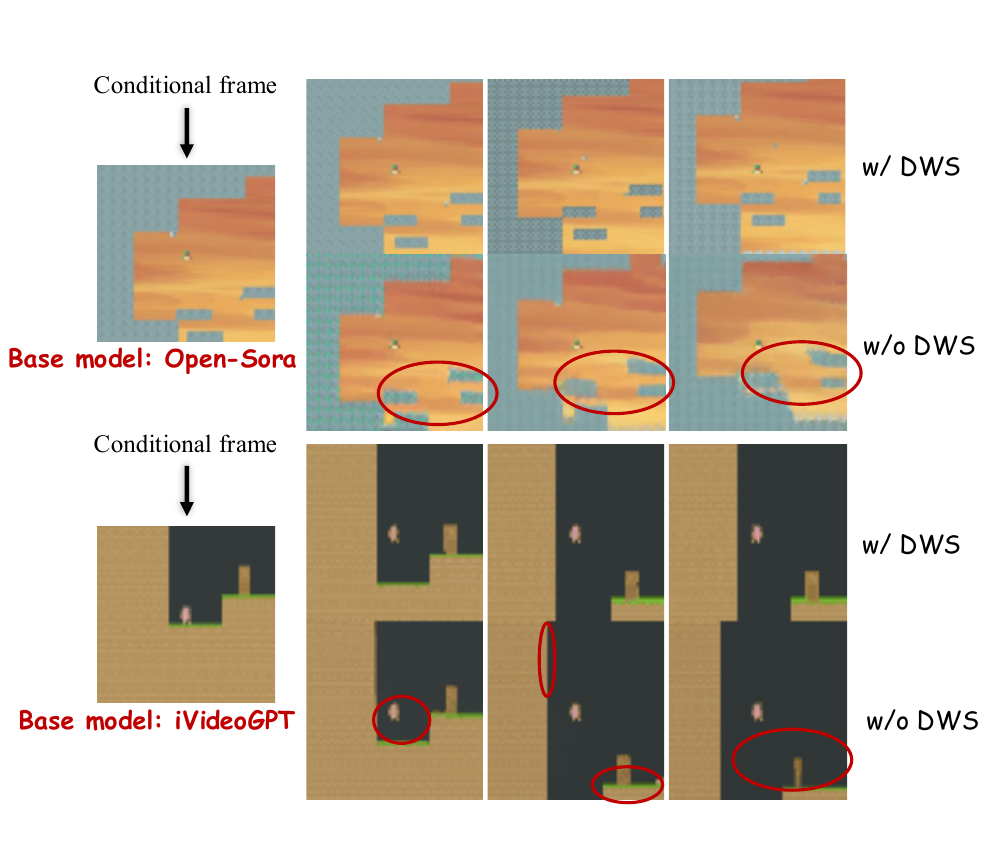}
    \vspace{-2.em}
    \caption{Quantitative comparison between pre-trained video generative models and fine-tuned models by our proposed methods. Base models without DWS can generate inconsistent pixels and fail to match the conditioned actions, as highlighted in red circles.}
    \vspace{-1.em}
    \label{fig:demo_comp}
\end{wrapfigure}

\textbf{Qualitative Results Analysis.} We qualitatively evaluate two different models, i.e., Open-Sora and iVideoGPT, fine-tuned by DWS using unseen initial frames. Fig.~\ref{fig:demos} showcases examples of video generations across diverse domains given unseen inputs. We observe that both base models fine-tuned by DWS successfully generate high-quality, controllable videos characterized by coherent temporal dynamics and consistent background preservation. Furthermore, the models demonstrate robust generalization capabilities when processing unseen inputs with varying backgrounds and textures, validating the effectiveness of our approach. As evidenced in Figure \ref{fig:demo_comp}, while the base models tend to generate videos with visual distortions, DWS significantly enhances the output quality by maintaining object consistency and producing precise visual predictions.
\textbf{Quantitative Results Analysis.} From the results shown in Table~\ref{tab:video_pred} and Table~\ref{tab:game_video_pred}, we
have the following key observations: (\romannumeral1) DWS demonstrates superior performance in action-conditioned video prediction across different base models. On the BAIR dataset with a continuous action space, DWS significantly enhances the performance of both the diffusion-based Open-Sora and the autoregressive-based iVideoGPT, highlighting its generalizability and versatility across different architectures. The results show that Open-Sora, a traditional text-to-video model that conditions generation on static, global text prompts, can gain significant improvements in action controllability after DWS fine-tuning. Specifically, we observe an 11.7\% reduction in FVD and a 27.9\% decrease in LPIPS. Even with iVideoGPT, which is specifically designed for action-conditioned video generation, DWS achieves notable performance improvements. Regarding both Procgen and Atari game datasets with discrete action spaces, DWS consistently improves the base models' performance by a distinct margin, yielding significant enhancements in generated video quality. These enhancements can be attributed to two key components: The action-conditioned module, which efficiently modulates each reaction to its corresponding frame, and the motion-reinforced loss function, which effectively captures frame-to-frame pixel dynamics. (\romannumeral2) DWS demonstrates its versatility as a universal method that can be efficiently deployed across diverse datasets, ranging from robotics datasets with continuous action spaces to game datasets with discrete action spaces. Furthermore, DWS can be seamlessly integrated into various architectures, where the two popular architectures considered in this work are diffusion-based and autoregressive transformer-based models.
\vspace{-0.5em}
\subsection{Model-Based Reinforcement Learning} 
To validate the practical utility of DWS-fine-tuned models for world simulators, we evaluate their performance by utilizing them as world models in MBRL policy learning.
\begin{figure*}[t]
     \centering
      \subfigure[Atari]{
      \centering
        \includegraphics[width=0.48\linewidth]{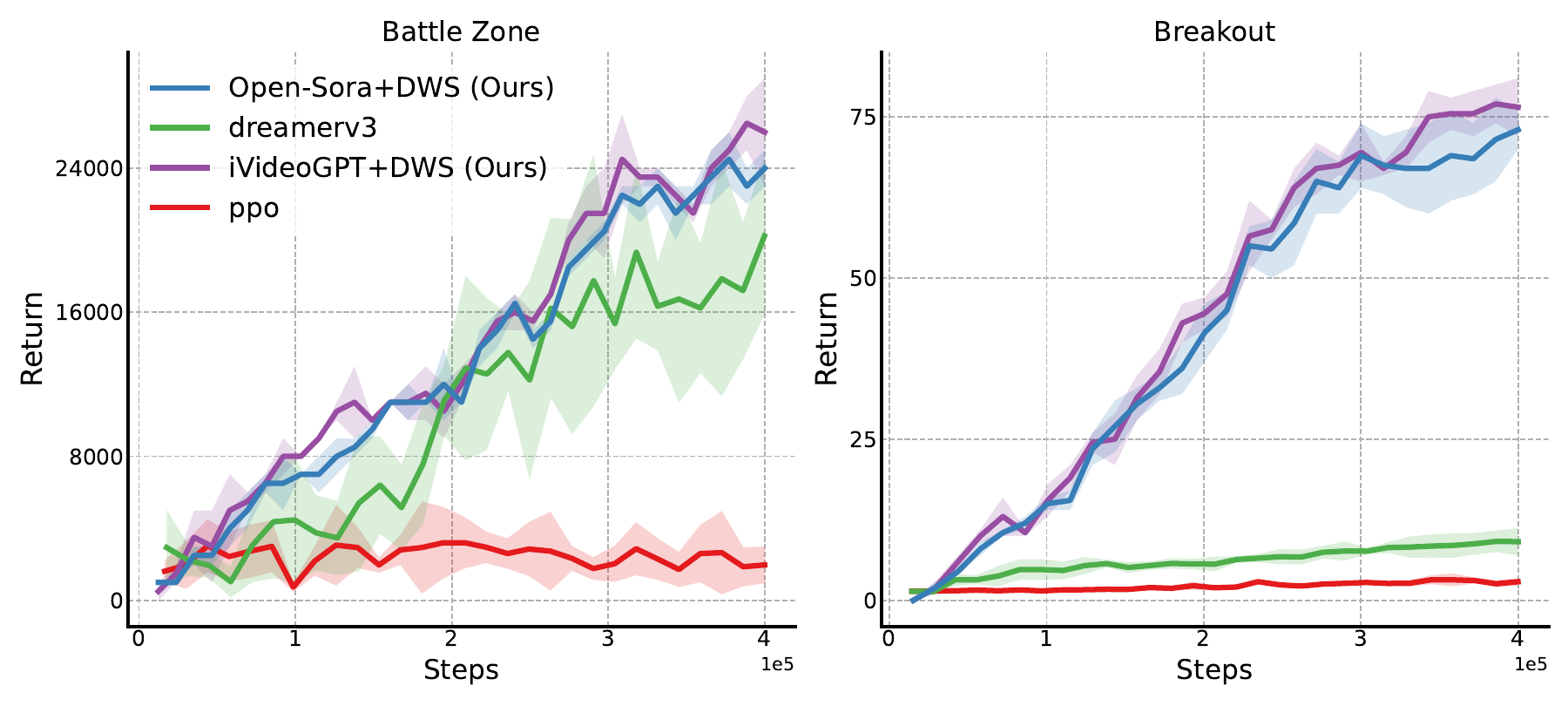}
    }\hspace{-2.3mm}
    \subfigure[Procgen]{
    \centering
        \includegraphics[width=0.48\linewidth]{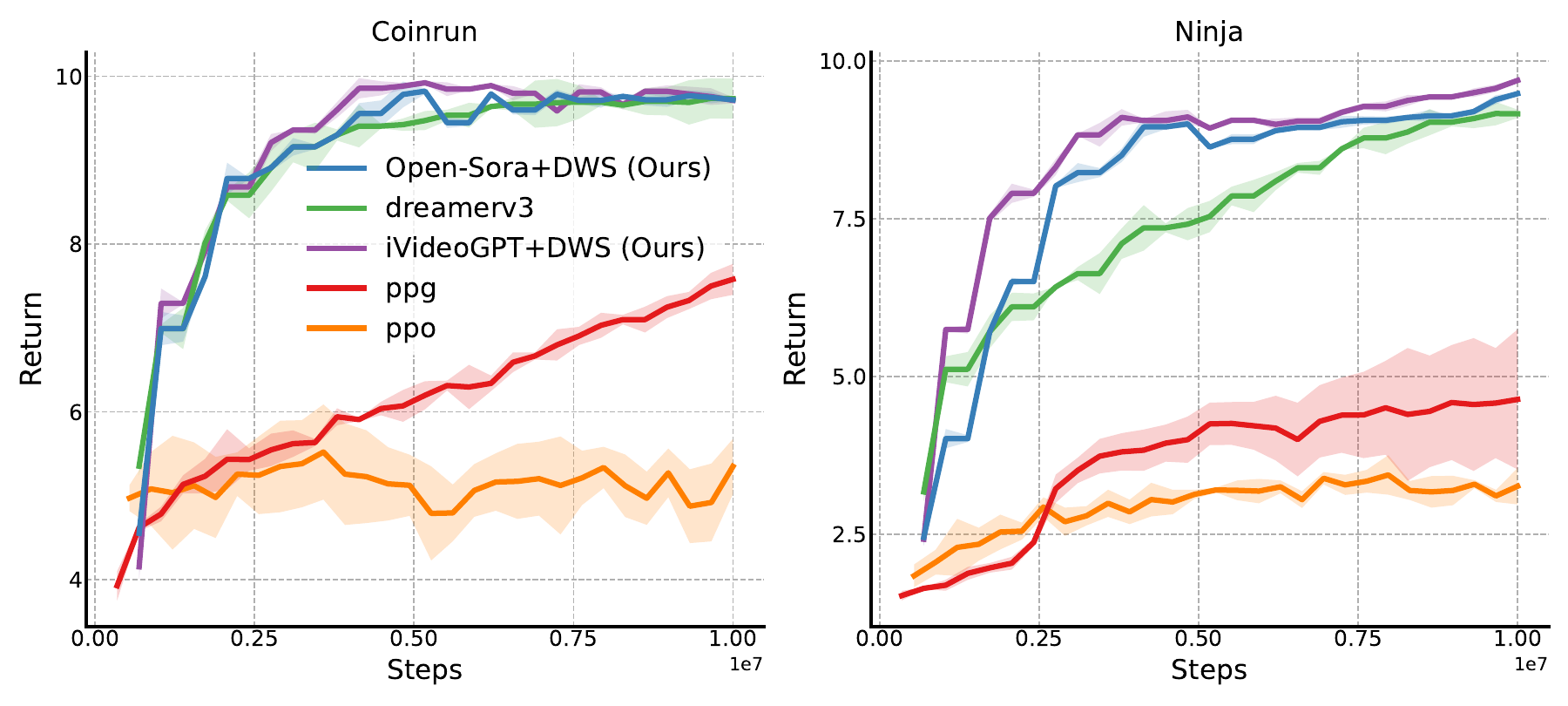}
    }
    \vspace{-.5em}
    \caption{Averaged Return across five random seeds on Atari and Procgen environments.}
     \label{fig:mbrl}
     \vspace{-1.2em}
 \end{figure*}
 
\textbf{Benchmarks and Baselines.} We conduct experiments on coinrun and ninja platformer games from the Procgen benchmark, and Breakout and Battle Zone games from the Atari benchmark. These environments take RBG images as observations and discrete actions for control. Procgen employs a 15-dimensional action space, Breakout operates with a 4-dimensional space, while Battle Zone utilizes an 18-dimensional action space. We compare our MBRL method with prioritized imagination with the following baselines: 1) PPO~\citep{schulman2017proximal} is a model-free RL method that is widely used. Our method is built on PPO. 2) Dreamerv3~\citep{hafner2023dreamerv3} is a SOTA model-based RL method that employs a recurrent network for dynamic prediction and actor-critic RL for policy learning, which is effective in handling tasks with discrete action spaces. For Procgen environments, we additionally include PPG~\citep{cobbe2021phasic} for comparison, as it represents a competitive algorithm on Procgen.
 
\begin{wrapfigure}{r}{0.55\linewidth}
     \centering
      \vspace{-.5em}
     \includegraphics[width=1.\linewidth]{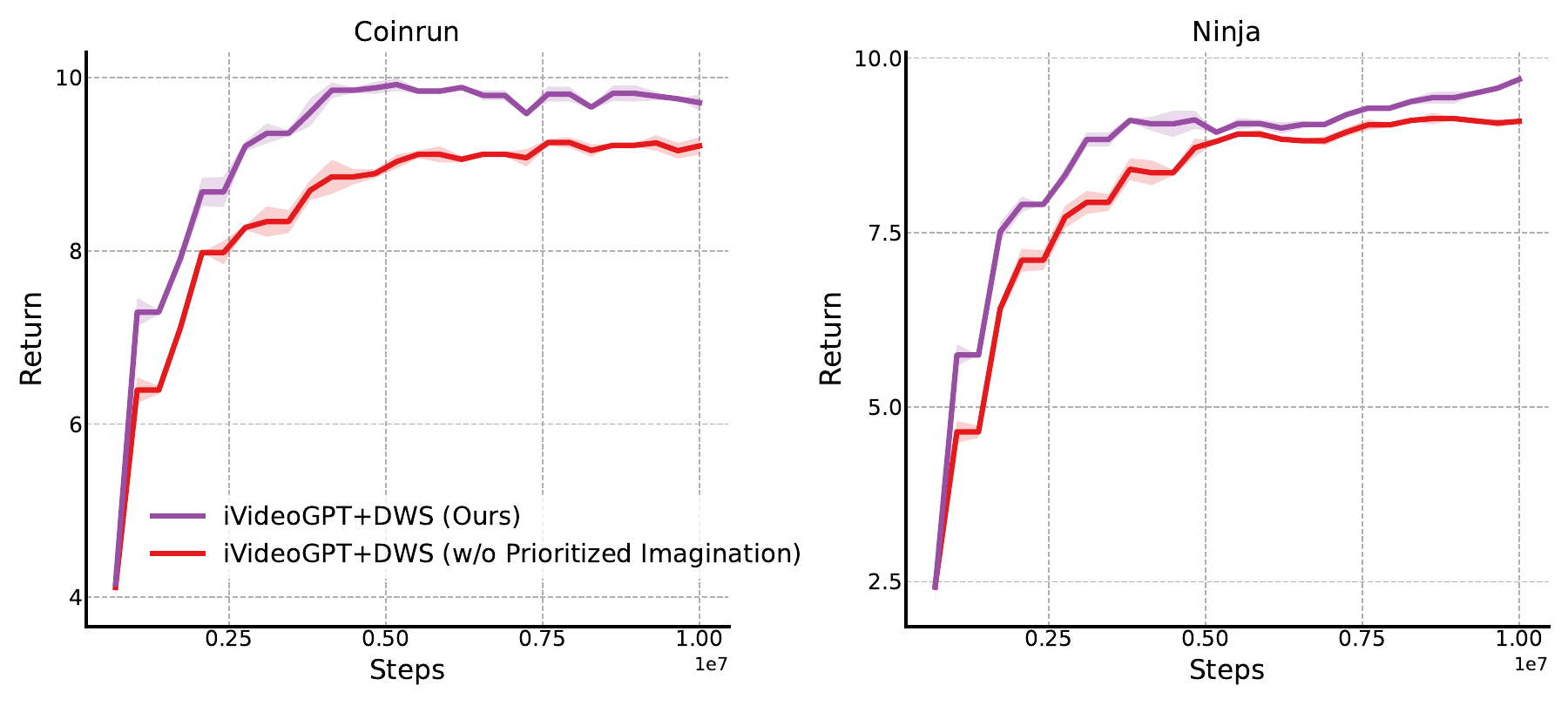}
     \vspace{-1.5em}
     \caption{Prioritized imagination improves the performance of model-based RL.}
     \vspace{-.8em}
     \label{fig:prioritized}
 \end{wrapfigure}

\textbf{Results Analysis.} The experimental results presented in Fig.~\ref{fig:mbrl} demonstrate that our DWS-trained world model, when combined with a simple PPO algorithm, significantly outperforms both vanilla PPO and state-of-the-art model-based reinforcement learning methods, i.e., Dreamerv3. In Procgen environments, DWS exhibits substantial performance improvements over model-free approaches such as PPO and PPG. Although the DWS-trained world model requires fine-tuning during MBRL policy training—due to the episodic variations in background and object details inherent to Procgen, it maintains competitive performance compared to existing model-based methods. In Atari environments, DWS demonstrates substantial performance improvements over existing methods, attributed to its world models having acquired comprehensive dynamics knowledge for action simulation. Specifically, in Breakout, DWS achieves a remarkable $7\times$ performance gain compared to SOTA methods. This superior performance, particularly in terms of sample efficiency, can be attributed to our proposed prioritized imagination technique. We validate this contribution through ablation studies conducted on Procgen environments, with results presented in Figure~\ref{fig:prioritized}.

\subsection{Offline Model-Based Reinforcement Learning}
We further explore the potential of leveraging trained video world models to augment offline datasets for policy enhancement. Using CoinRun and Ninja environments as case studies, we first establish baseline datasets by collecting 1M expert trajectories for each environment using a well-trained PPO agent, following \citet{mediratta2024the}. For each evaluated environment, we employ \textit{Open-Sora+DWS} to synthesize an additional 1M state-action transitions during training, effectively doubling the size of the original datasets. To validate the effectiveness of this data augmentation approach, we evaluate the performance improvements using two different offline RL algorithms: Conservative Q-Learning (CQL)~\citep{kumar2020cql} and Implicit Q-Learning (IQL)~\citep{kostrikov2022iql}. As shown in Table~\ref{table:augmentation}, augmenting offline RL with world model-generated data during training significantly enhances performance across different algorithms and environments. These results demonstrate that the DWS-trained world simulator can generate meaningful state-action-reward transitions that effectively supplement the offline dataset for policy learning. The success of this approach highlights the potential of using pre-trained video models as world simulators for reinforcement learning applications. Furthermore, these results validate that DWS-trained models can generate accurate dynamic transitions, making them valuable tools for policy learning.
\begin{table}[htbp]
\begin{center}
\vspace{-1.em}
\caption{Average return across 3 seeds on Coinrun and Ninja tasks.}
\label{table:augmentation}
\resizebox{0.6\linewidth}{!}{
\begin{tabular}{ccc|cc} 
    \hline
    \textbf{Tasks}&\textbf{CQL}&\textbf{CQL w/ wm}&\textbf{IQL}&\textbf{IQL w/ wm}\\
    \hline
    \textbf{Coinrun}&$8.58\pm0.29$ &$\bf{8.81\pm0.21}\color{blue}(\uparrow)$ &$8.52\pm0.26$ &$\bf{8.93\pm0.19}\color{blue}(\uparrow)$ \\
     \textbf{Ninja}&$5.92\pm0.2$ &$\bf{6.31\pm0.23}\color{blue}(\uparrow)$ &$5.7\pm0.35$ &$\bf{6.33\pm0.17}\color{blue}(\uparrow)$\\
      \hline
    \end{tabular}
    \vspace{-1.em}
    }
\vspace{-1em}
\end{center}
\end{table}
\vspace{-1em}
\section{Conclusion and Limitation}
In this paper, we present DWS, a novel approach that efficiently adapts pre-trained video generative models as world simulators by leveraging their rich prior knowledge learned from large-scale datasets for downstream action-conditioned simulation. Our framework introduces a lightweight action-conditioned module that enables action-frame alignment and can be seamlessly integrated into various model architectures, and a motion-reinforced loss specifically designed to model inter-frame pixel dynamics crucial for accurate world simulation. Extensive experimental results demonstrate that DWS significantly improves both video prediction quality and MBRL performance, leading to meaningful applications. However, DWS currently is limited in modeling videos with extended temporal horizons or high spatial resolutions. We leave it as future work.

\bibliography{main}

\begin{thebibliography}{68}
\providecommand{\natexlab}[1]{#1}
\providecommand{\url}[1]{\texttt{#1}}
\expandafter\ifx\csname urlstyle\endcsname\relax
  \providecommand{\doi}[1]{doi: #1}\else
  \providecommand{\doi}{doi: \begingroup \urlstyle{rm}\Url}\fi

\bibitem[Alonso et~al.(2024)Alonso, Jelley, Micheli, Kanervisto, Storkey, Pearce, and Fleuret]{alonso2024diamond}
Eloi Alonso, Adam Jelley, Vincent Micheli, Anssi Kanervisto, Amos Storkey, Tim Pearce, and François Fleuret.
\newblock Diffusion for world modeling: Visual details matter in atari.
\newblock In \emph{Thirty-eighth Conference on Neural Information Processing Systems}, 2024.
\newblock URL \url{https://arxiv.org/abs/2405.12399}.

\bibitem[Babaeizadeh et~al.(2021)Babaeizadeh, Saffar, Nair, Levine, Finn, and Erhan]{babaeizadeh2021fitvid}
Mohammad Babaeizadeh, Mohammad~Taghi Saffar, Suraj Nair, Sergey Levine, Chelsea Finn, and Dumitru Erhan.
\newblock Fitvid: Overfitting in pixel-level video prediction.
\newblock \emph{arXiv preprint arXiv:2106.13195}, 2021.

\bibitem[Bain et~al.(2021)Bain, Nagrani, Varol, and Zisserman]{Bain21}
Max Bain, Arsha Nagrani, G{\"u}l Varol, and Andrew Zisserman.
\newblock Frozen in time: A joint video and image encoder for end-to-end retrieval.
\newblock In \emph{IEEE International Conference on Computer Vision}, 2021.

\bibitem[Bellemare et~al.(2013)Bellemare, Naddaf, Veness, and Bowling]{bellemare2013arcade}
Marc~G Bellemare, Yavar Naddaf, Joel Veness, and Michael Bowling.
\newblock The arcade learning environment: An evaluation platform for general agents.
\newblock \emph{Journal of Artificial Intelligence Research}, 47:\penalty0 253--279, 2013.

\bibitem[Blattmann et~al.(2023)Blattmann, Dockhorn, Kulal, Mendelevitch, Kilian, Lorenz, Levi, English, Voleti, Letts, et~al.]{blattmann2023SDVideo}
Andreas Blattmann, Tim Dockhorn, Sumith Kulal, Daniel Mendelevitch, Maciej Kilian, Dominik Lorenz, Yam Levi, Zion English, Vikram Voleti, Adam Letts, et~al.
\newblock Stable video diffusion: Scaling latent video diffusion models to large datasets.
\newblock \emph{arXiv preprint arXiv:2311.15127}, 2023.

\bibitem[Brooks et~al.(2024)Brooks, Peebles, Holmes, DePue, Guo, Jing, Schnurr, Taylor, Luhman, Luhman, Ng, Wang, and Ramesh]{videoworldsimulators2024}
Tim Brooks, Bill Peebles, Connor Holmes, Will DePue, Yufei Guo, Li~Jing, David Schnurr, Joe Taylor, Troy Luhman, Eric Luhman, Clarence Ng, Ricky Wang, and Aditya Ramesh.
\newblock Video generation models as world simulators.
\newblock 2024.

\bibitem[Bruce et~al.(2024)Bruce, Dennis, Edwards, Parker-Holder, Shi, Hughes, Lai, Mavalankar, Steigerwald, Apps, Aytar, Bechtle, Behbahani, Chan, Heess, Gonzalez, Osindero, Ozair, Reed, Zhang, Zolna, Clune, de~Freitas, Singh, and Rockt{\"a}schel]{bruce2024genie}
Jake Bruce, Michael~D Dennis, Ashley Edwards, Jack Parker-Holder, Yuge Shi, Edward Hughes, Matthew Lai, Aditi Mavalankar, Richie Steigerwald, Chris Apps, Yusuf Aytar, Sarah Maria~Elisabeth Bechtle, Feryal Behbahani, Stephanie~C.Y. Chan, Nicolas Heess, Lucy Gonzalez, Simon Osindero, Sherjil Ozair, Scott Reed, Jingwei Zhang, Konrad Zolna, Jeff Clune, Nando de~Freitas, Satinder Singh, and Tim Rockt{\"a}schel.
\newblock Genie: Generative interactive environments.
\newblock In \emph{Forty-first International Conference on Machine Learning}, 2024.

\bibitem[Carreira and Zisserman(2017)]{carreira2017i3d}
Joao Carreira and Andrew Zisserman.
\newblock Quo vadis, action recognition? a new model and the kinetics dataset.
\newblock In \emph{proceedings of the IEEE Conference on Computer Vision and Pattern Recognition}, pages 6299--6308, 2017.

\bibitem[Che et~al.(2025)Che, He, Liu, Jin, and Chen]{2025gamegen}
Haoxuan Che, Xuanhua He, Quande Liu, Cheng Jin, and Hao Chen.
\newblock Gamegen-\${\textbackslash}mathbb\{X\}\$: Interactive open-world game video generation.
\newblock In \emph{The Thirteenth International Conference on Learning Representations}, 2025.
\newblock URL \url{https://openreview.net/forum?id=8VG8tpPZhe}.

\bibitem[Chen et~al.(2023)Chen, Yu, Ge, Yao, Xie, Wu, Wang, Kwok, Luo, Lu, and Li]{chen2023pixartalpha}
Junsong Chen, Jincheng Yu, Chongjian Ge, Lewei Yao, Enze Xie, Yue Wu, Zhongdao Wang, James Kwok, Ping Luo, Huchuan Lu, and Zhenguo Li.
\newblock Pixart-$\alpha$: Fast training of diffusion transformer for photorealistic text-to-image synthesis, 2023.

\bibitem[Chen et~al.(2024)Chen, Siarohin, Menapace, Deyneka, Chao, Jeon, Fang, Lee, Ren, Yang, and Tulyakov]{chen2024panda70m}
Tsai-Shien Chen, Aliaksandr Siarohin, Willi Menapace, Ekaterina Deyneka, Hsiang-wei Chao, Byung~Eun Jeon, Yuwei Fang, Hsin-Ying Lee, Jian Ren, Ming-Hsuan Yang, and Sergey Tulyakov.
\newblock Panda-70m: Captioning 70m videos with multiple cross-modality teachers.
\newblock In \emph{Proceedings of the IEEE/CVF Conference on Computer Vision and Pattern Recognition}, 2024.

\bibitem[Cobbe et~al.(2020)Cobbe, Hesse, Hilton, and Schulman]{cobbe2020leveraging}
Karl Cobbe, Chris Hesse, Jacob Hilton, and John Schulman.
\newblock Leveraging procedural generation to benchmark reinforcement learning.
\newblock In \emph{International conference on machine learning}, pages 2048--2056. PMLR, 2020.

\bibitem[Cobbe et~al.(2021)Cobbe, Hilton, Klimov, and Schulman]{cobbe2021phasic}
Karl~W Cobbe, Jacob Hilton, Oleg Klimov, and John Schulman.
\newblock Phasic policy gradient.
\newblock In \emph{International Conference on Machine Learning}, pages 2020--2027. PMLR, 2021.

\bibitem[Decart et~al.(2024)Decart, Quevedo, McIntyre, Campbell, Chen, and Wachen]{oasis2024}
Decart, Julian Quevedo, Quinn McIntyre, Spruce Campbell, Xinlei Chen, and Robert Wachen.
\newblock Oasis: A universe in a transformer.
\newblock 2024.
\newblock URL \url{https://oasis-model.github.io/}.

\bibitem[Ding et~al.(2024)Ding, Zhang, Tian, and Zheng]{ding2024dwm}
Zihan Ding, Amy Zhang, Yuandong Tian, and Qinqing Zheng.
\newblock Diffusion world model.
\newblock \emph{arXiv preprint arXiv:2402.03570}, 2024.

\bibitem[Ebert et~al.(2017)Ebert, Finn, Lee, and Levine]{ebert2017self}
Frederik Ebert, Chelsea Finn, Alex~X Lee, and Sergey Levine.
\newblock Self-supervised visual planning with temporal skip connections.
\newblock \emph{CoRL}, 12\penalty0 (16):\penalty0 23, 2017.

\bibitem[Feng et~al.(2024)Feng, Zhang, Yang, Xiao, Shu, Liu, Zheng, Huang, Liu, and Zhang]{feng2024matrix}
Ruili Feng, Han Zhang, Zhantao Yang, Jie Xiao, Zhilei Shu, Zhiheng Liu, Andy Zheng, Yukun Huang, Yu~Liu, and Hongyang Zhang.
\newblock The matrix: Infinite-horizon world generation with real-time moving control.
\newblock \emph{arXiv preprint arXiv:2412.03568}, 2024.

\bibitem[Gupta et~al.(2022)Gupta, Tian, Zhang, Wu, Mart{\'\i}n-Mart{\'\i}n, and Fei-Fei]{gupta2022maskvit}
Agrim Gupta, Stephen Tian, Yunzhi Zhang, Jiajun Wu, Roberto Mart{\'\i}n-Mart{\'\i}n, and Li~Fei-Fei.
\newblock Maskvit: Masked visual pre-training for video prediction.
\newblock \emph{arXiv preprint arXiv:2206.11894}, 2022.

\bibitem[Hafner et~al.(2019)Hafner, Lillicrap, Fischer, Villegas, Ha, Lee, and Davidson]{hafner2019PlaNet}
Danijar Hafner, Timothy Lillicrap, Ian Fischer, Ruben Villegas, David Ha, Honglak Lee, and James Davidson.
\newblock Learning latent dynamics for planning from pixels.
\newblock In \emph{International conference on machine learning}, pages 2555--2565. PMLR, 2019.

\bibitem[Hafner et~al.(2020{\natexlab{a}})Hafner, Lillicrap, Ba, and Norouzi]{Hafner2020Dream}
Danijar Hafner, Timothy Lillicrap, Jimmy Ba, and Mohammad Norouzi.
\newblock Dream to control: Learning behaviors by latent imagination.
\newblock In \emph{International Conference on Learning Representations}, 2020{\natexlab{a}}.
\newblock URL \url{https://openreview.net/forum?id=S1lOTC4tDS}.

\bibitem[Hafner et~al.(2020{\natexlab{b}})Hafner, Lillicrap, Norouzi, and Ba]{hafner2020dreamerv2}
Danijar Hafner, Timothy Lillicrap, Mohammad Norouzi, and Jimmy Ba.
\newblock Mastering atari with discrete world models.
\newblock \emph{arXiv preprint arXiv:2010.02193}, 2020{\natexlab{b}}.

\bibitem[Hafner et~al.(2023)Hafner, Pasukonis, Ba, and Lillicrap]{hafner2023dreamerv3}
Danijar Hafner, Jurgis Pasukonis, Jimmy Ba, and Timothy Lillicrap.
\newblock Mastering diverse domains through world models.
\newblock \emph{arXiv preprint arXiv:2301.04104}, 2023.

\bibitem[Ho et~al.(2020)Ho, Jain, and Abbeel]{ddpm}
Jonathan Ho, Ajay Jain, and Pieter Abbeel.
\newblock Denoising diffusion probabilistic models.
\newblock \emph{Advances in neural information processing systems}, 33:\penalty0 6840--6851, 2020.

\bibitem[Huang et~al.(2022)Huang, Dossa, Raffin, Kanervisto, and Wang]{shengyi2022the37implementation}
Shengyi Huang, Rousslan Fernand~Julien Dossa, Antonin Raffin, Anssi Kanervisto, and Weixun Wang.
\newblock The 37 implementation details of proximal policy optimization.
\newblock In \emph{ICLR Blog Track}, 2022.
\newblock URL \url{https://iclr-blog-track.github.io/2022/03/25/ppo-implementation-details/}.
\newblock https://iclr-blog-track.github.io/2022/03/25/ppo-implementation-details/.

\bibitem[Huynh-Thu and Ghanbari(2008)]{HuynhThu2008psnr}
Quan Huynh-Thu and Mohammed Ghanbari.
\newblock Scope of validity of psnr in image/video quality assessment.
\newblock \emph{Electronics Letters}, 44:\penalty0 800--801, 2008.
\newblock URL \url{https://api.semanticscholar.org/CorpusID:62732555}.

\bibitem[Janner et~al.(2019{\natexlab{a}})Janner, Fu, Zhang, and Levine]{janner2019mbpo}
Michael Janner, Justin Fu, Marvin Zhang, and Sergey Levine.
\newblock When to trust your model: Model-based policy optimization.
\newblock \emph{Advances in neural information processing systems}, 32, 2019{\natexlab{a}}.

\bibitem[Janner et~al.(2019{\natexlab{b}})Janner, Fu, Zhang, and Levine]{janner2019trust}
Michael Janner, Justin Fu, Marvin Zhang, and Sergey Levine.
\newblock When to trust your model: Model-based policy optimization.
\newblock \emph{Advances in neural information processing systems}, 32, 2019{\natexlab{b}}.

\bibitem[Kostrikov et~al.(2022)Kostrikov, Nair, and Levine]{kostrikov2022iql}
Ilya Kostrikov, Ashvin Nair, and Sergey Levine.
\newblock Offline reinforcement learning with implicit q-learning.
\newblock In \emph{International Conference on Learning Representations}, 2022.
\newblock URL \url{https://openreview.net/forum?id=68n2s9ZJWF8}.

\bibitem[Kumar et~al.(2020)Kumar, Zhou, Tucker, and Levine]{kumar2020cql}
Aviral Kumar, Aurick Zhou, George Tucker, and Sergey Levine.
\newblock Conservative q-learning for offline reinforcement learning.
\newblock \emph{Advances in Neural Information Processing Systems}, 33:\penalty0 1179--1191, 2020.

\bibitem[Lin et~al.(2024)Lin, Ge, Cheng, Li, Zhu, Wang, He, Ye, Yuan, Chen, et~al.]{lin2024opensoraplan}
Bin Lin, Yunyang Ge, Xinhua Cheng, Zongjian Li, Bin Zhu, Shaodong Wang, Xianyi He, Yang Ye, Shenghai Yuan, Liuhan Chen, et~al.
\newblock Open-sora plan: Open-source large video generation model.
\newblock \emph{arXiv preprint arXiv:2412.00131}, 2024.

\bibitem[Liu et~al.(2022)Liu, Gong, and Liu]{liu2022flow}
Xingchao Liu, Chengyue Gong, and Qiang Liu.
\newblock Flow straight and fast: Learning to generate and transfer data with rectified flow.
\newblock \emph{arXiv preprint arXiv:2209.03003}, 2022.

\bibitem[Ma et~al.(2024)Ma, Wang, Jia, Chen, Liu, Li, Chen, and Qiao]{ma2024latte}
Xin Ma, Yaohui Wang, Gengyun Jia, Xinyuan Chen, Ziwei Liu, Yuan-Fang Li, Cunjian Chen, and Yu~Qiao.
\newblock Latte: Latent diffusion transformer for video generation.
\newblock \emph{arXiv preprint arXiv:2401.03048}, 2024.

\bibitem[Machado et~al.(2018)Machado, Bellemare, Talvitie, Veness, Hausknecht, and Bowling]{atari}
Marlos~C Machado, Marc~G Bellemare, Erik Talvitie, Joel Veness, Matthew Hausknecht, and Michael Bowling.
\newblock Revisiting the arcade learning environment: Evaluation protocols and open problems for general agents.
\newblock \emph{Journal of Artificial Intelligence Research}, 61:\penalty0 523--562, 2018.

\bibitem[Mediratta et~al.(2024)Mediratta, You, Jiang, and Raileanu]{mediratta2024the}
Ishita Mediratta, Qingfei You, Minqi Jiang, and Roberta Raileanu.
\newblock The generalization gap in offline reinforcement learning.
\newblock In \emph{The Twelfth International Conference on Learning Representations}, 2024.
\newblock URL \url{https://openreview.net/forum?id=3w6xuXDOdY}.

\bibitem[Micheli et~al.(2023{\natexlab{a}})Micheli, Alonso, and Fleuret]{micheli2023iris}
Vincent Micheli, Eloi Alonso, and Fran{\c{c}}ois Fleuret.
\newblock Transformers are sample-efficient world models.
\newblock In \emph{The Eleventh International Conference on Learning Representations}, 2023{\natexlab{a}}.
\newblock URL \url{https://openreview.net/forum?id=vhFu1Acb0xb}.

\bibitem[Micheli et~al.(2023{\natexlab{b}})Micheli, Alonso, and Fleuret]{micheli2023transformers}
Vincent Micheli, Eloi Alonso, and Fran{\c{c}}ois Fleuret.
\newblock Transformers are sample-efficient world models.
\newblock In \emph{The Eleventh International Conference on Learning Representations}, 2023{\natexlab{b}}.
\newblock URL \url{https://openreview.net/forum?id=vhFu1Acb0xb}.

\bibitem[Mnih et~al.(2013)Mnih, Kavukcuoglu, Silver, Graves, Antonoglou, Wierstra, and Riedmiller]{mnih2013dqn}
Volodymyr Mnih, Koray Kavukcuoglu, David Silver, Alex Graves, Ioannis Antonoglou, Daan Wierstra, and Martin Riedmiller.
\newblock Playing atari with deep reinforcement learning.
\newblock \emph{arXiv preprint arXiv:1312.5602}, 2013.

\bibitem[Parker-Holder et~al.(2024)Parker-Holder, Ball, Bruce, Dasagi, Holsheimer, Kaplanis, Moufarek, Scully, Shar, Shi, Spencer, Yung, Dennis, Kenjeyev, Long, Mnih, Chan, Gazeau, Li, Pardo, Wang, Zhang, Besse, Harley, Mitenkova, Wang, Clune, Hassabis, Hadsell, Bolton, Singh, and Rocktäschel]{genie2}
Jack Parker-Holder, Philip Ball, Jake Bruce, Vibhavari Dasagi, Kristian Holsheimer, Christos Kaplanis, Alexandre Moufarek, Guy Scully, Jeremy Shar, Jimmy Shi, Stephen Spencer, Jessica Yung, Michael Dennis, Sultan Kenjeyev, Shangbang Long, Vlad Mnih, Harris Chan, Maxime Gazeau, Bonnie Li, Fabio Pardo, Luyu Wang, Lei Zhang, Frederic Besse, Tim Harley, Anna Mitenkova, Jane Wang, Jeff Clune, Demis Hassabis, Raia Hadsell, Adrian Bolton, Satinder Singh, and Tim Rocktäschel.
\newblock Genie 2: A large-scale foundation world model.
\newblock 2024.

\bibitem[Peebles and Xie(2023)]{peebles2023dit}
William Peebles and Saining Xie.
\newblock Scalable diffusion models with transformers.
\newblock In \emph{Proceedings of the IEEE/CVF International Conference on Computer Vision}, pages 4195--4205, 2023.

\bibitem[Polyak et~al.(2024)Polyak, Zohar, Brown, Tjandra, Sinha, Lee, Vyas, Shi, Ma, Chuang, et~al.]{polyak2024movie}
Adam Polyak, Amit Zohar, Andrew Brown, Andros Tjandra, Animesh Sinha, Ann Lee, Apoorv Vyas, Bowen Shi, Chih-Yao Ma, Ching-Yao Chuang, et~al.
\newblock Movie gen: A cast of media foundation models.
\newblock \emph{arXiv preprint arXiv:2410.13720}, 2024.

\bibitem[Rigter et~al.(2024)Rigter, Gupta, Hilmkil, and Ma]{rigter2024avid}
Marc Rigter, Tarun Gupta, Agrin Hilmkil, and Chao Ma.
\newblock Avid: Adapting video diffusion models to world models.
\newblock \emph{arXiv preprint arXiv:2410.12822}, 2024.

\bibitem[Robine et~al.(2023)Robine, H{\"o}ftmann, Uelwer, and Harmeling]{robine2023twm}
Jan Robine, Marc H{\"o}ftmann, Tobias Uelwer, and Stefan Harmeling.
\newblock Transformer-based world models are happy with 100k interactions.
\newblock In \emph{The Eleventh International Conference on Learning Representations}, 2023.
\newblock URL \url{https://openreview.net/forum?id=TdBaDGCpjly}.

\bibitem[Schulman et~al.(2017)Schulman, Wolski, Dhariwal, Radford, and Klimov]{schulman2017proximal}
John Schulman, Filip Wolski, Prafulla Dhariwal, Alec Radford, and Oleg Klimov.
\newblock Proximal policy optimization algorithms.
\newblock \emph{arXiv preprint arXiv:1707.06347}, 2017.

\bibitem[Sharma et~al.(2024)Sharma, Yu, Razavi, Toor, Pierson, Gupta, Waters, van~den Oord, Tanis, Erhan, Lau, Shaw, Barth-Maron, Shaw, Zhang, Nandwani, Moraldo, Kim, Blok, Bauer, Donahue, Chung, Mathewson, David, Espeholt, van Zee, McGill, Narasimhan, Wang, Bińkowski, Babaeizadeh, Saffar, de~Freitas, Pezzotti, Kindermans, Rane, Hornung, Riachi, Villegas, Qian, Dieleman, Zhang, Cabi, Luo, Fruchter, Nørly, Srinivasan, Pfaff, Hume, Verma, Hua, Zhu, Yan, Wang, Kim, Du, and Chen]{veo2024}
Abhishek Sharma, Adams Yu, Ali Razavi, Andeep Toor, Andrew Pierson, Ankush Gupta, Austin Waters, Aäron van~den Oord, Daniel Tanis, Dumitru Erhan, Eric Lau, Eleni Shaw, Gabe Barth-Maron, Greg Shaw, Han Zhang, Henna Nandwani, Hernan Moraldo, Hyunjik Kim, Irina Blok, Jakob Bauer, Jeff Donahue, Junyoung Chung, Kory Mathewson, Kurtis David, Lasse Espeholt, Marc van Zee, Matt McGill, Medhini Narasimhan, Miaosen Wang, Mikołaj Bińkowski, Mohammad Babaeizadeh, Mohammad~Taghi Saffar, Nando de~Freitas, Nick Pezzotti, Pieter-Jan Kindermans, Poorva Rane, Rachel Hornung, Robert Riachi, Ruben Villegas, Rui Qian, Sander Dieleman, Serena Zhang, Serkan Cabi, Shixin Luo, Shlomi Fruchter, Signe Nørly, Srivatsan Srinivasan, Tobias Pfaff, Tom Hume, Vikas Verma, Weizhe Hua, William Zhu, Xinchen Yan, Xinyu Wang, Yelin Kim, Yuqing Du, and Yutian Chen.
\newblock Veo.
\newblock 2024.
\newblock URL \url{https://deepmind.google/technologies/veo/}.

\bibitem[Simonyan and Zisserman(2014)]{simonyan2014vgg}
Karen Simonyan and Andrew Zisserman.
\newblock Very deep convolutional networks for large-scale image recognition.
\newblock \emph{arXiv preprint arXiv:1409.1556}, 2014.

\bibitem[Sutton(1991)]{sutton1991dyna}
Richard~S Sutton.
\newblock Dyna, an integrated architecture for learning, planning, and reacting.
\newblock \emph{ACM Sigart Bulletin}, 2\penalty0 (4):\penalty0 160--163, 1991.

\bibitem[Sutton and Barto(1998)]{Sutton1998ReinforcementL}
Richard~S. Sutton and Andrew~G. Barto.
\newblock Reinforcement learning - an introduction.
\newblock In \emph{Adaptive computation and machine learning}, 1998.
\newblock URL \url{https://api.semanticscholar.org/CorpusID:264703640}.

\bibitem[Tassa et~al.(2018)Tassa, Doron, Muldal, Erez, Li, Casas, Budden, Abdolmaleki, Merel, Lefrancq, et~al.]{tassa2018deepmind}
Yuval Tassa, Yotam Doron, Alistair Muldal, Tom Erez, Yazhe Li, Diego de~Las Casas, David Budden, Abbas Abdolmaleki, Josh Merel, Andrew Lefrancq, et~al.
\newblock Deepmind control suite.
\newblock \emph{arXiv preprint arXiv:1801.00690}, 2018.

\bibitem[Tian et~al.(2024)Tian, Jiang, Yuan, PENG, and Wang]{tian2024visual}
Keyu Tian, Yi~Jiang, Zehuan Yuan, BINGYUE PENG, and Liwei Wang.
\newblock Visual autoregressive modeling: Scalable image generation via next-scale prediction.
\newblock In \emph{The Thirty-eighth Annual Conference on Neural Information Processing Systems}, 2024.
\newblock URL \url{https://openreview.net/forum?id=gojL67CfS8}.

\bibitem[Touvron et~al.(2023)Touvron, Lavril, Izacard, Martinet, Lachaux, Lacroix, Rozi{\`e}re, Goyal, Hambro, Azhar, et~al.]{touvron2023llama}
Hugo Touvron, Thibaut Lavril, Gautier Izacard, Xavier Martinet, Marie-Anne Lachaux, Timoth{\'e}e Lacroix, Baptiste Rozi{\`e}re, Naman Goyal, Eric Hambro, Faisal Azhar, et~al.
\newblock Llama: Open and efficient foundation language models.
\newblock \emph{arXiv preprint arXiv:2302.13971}, 2023.

\bibitem[Unterthiner et~al.(2018)Unterthiner, Van~Steenkiste, Kurach, Marinier, Michalski, and Gelly]{unterthiner2018fvd}
Thomas Unterthiner, Sjoerd Van~Steenkiste, Karol Kurach, Raphael Marinier, Marcin Michalski, and Sylvain Gelly.
\newblock Towards accurate generative models of video: A new metric \& challenges.
\newblock \emph{arXiv preprint arXiv:1812.01717}, 2018.

\bibitem[Valevski et~al.(2024)Valevski, Leviathan, Arar, and Fruchter]{valevski2024diffusion}
Dani Valevski, Yaniv Leviathan, Moab Arar, and Shlomi Fruchter.
\newblock Diffusion models are real-time game engines.
\newblock \emph{arXiv preprint arXiv:2408.14837}, 2024.

\bibitem[Vaswani et~al.(2017)Vaswani, Shazeer, Parmar, Uszkoreit, Jones, Gomez, Kaiser, and Polosukhin]{vaswani2017attention}
Ashish Vaswani, Noam Shazeer, Niki Parmar, Jakob Uszkoreit, Llion Jones, Aidan~N Gomez, {\L}ukasz Kaiser, and Illia Polosukhin.
\newblock Attention is all you need.
\newblock \emph{Advances in Neural Information Processing Systems}, 2017.

\bibitem[Voleti et~al.(2022)Voleti, Jolicoeur-Martineau, and Pal]{voleti2022mcvd}
Vikram Voleti, Alexia Jolicoeur-Martineau, and Chris Pal.
\newblock Mcvd-masked conditional video diffusion for prediction, generation, and interpolation.
\newblock \emph{Advances in neural information processing systems}, 35:\penalty0 23371--23385, 2022.

\bibitem[Wu et~al.(2024)Wu, Yin, Feng, He, Li, Hao, and Long]{wu2024ivideogpt}
Jialong Wu, Shaofeng Yin, Ningya Feng, Xu~He, Dong Li, Jianye Hao, and Mingsheng Long.
\newblock ivideogpt: Interactive videogpts are scalable world models.
\newblock In \emph{The Thirty-eighth Annual Conference on Neural Information Processing Systems}, 2024.

\bibitem[Xiang et~al.(2024)Xiang, Liu, Gu, Gao, Ning, Zha, Feng, Tao, Hao, Shi, et~al.]{xiang2024pandora}
Jiannan Xiang, Guangyi Liu, Yi~Gu, Qiyue Gao, Yuting Ning, Yuheng Zha, Zeyu Feng, Tianhua Tao, Shibo Hao, Yemin Shi, et~al.
\newblock Pandora: Towards general world model with natural language actions and video states.
\newblock \emph{arXiv preprint arXiv:2406.09455}, 2024.

\bibitem[Yan et~al.(2021)Yan, Zhang, Abbeel, and Srinivas]{yan2021videogpt}
Wilson Yan, Yunzhi Zhang, Pieter Abbeel, and Aravind Srinivas.
\newblock Videogpt: Video generation using vq-vae and transformers.
\newblock \emph{arXiv preprint arXiv:2104.10157}, 2021.

\bibitem[Yang et~al.(2023)Yang, Du, Ghasemipour, Tompson, Schuurmans, and Abbeel]{yang2023learning}
Mengjiao Yang, Yilun Du, Kamyar Ghasemipour, Jonathan Tompson, Dale Schuurmans, and Pieter Abbeel.
\newblock Learning interactive real-world simulators.
\newblock \emph{arXiv preprint arXiv:2310.06114}, 2023.

\bibitem[Yang et~al.(2024{\natexlab{a}})Yang, Li, Fang, Chen, Yu, Fu, Yang, and Ye]{yang2024playable}
Mingyu Yang, Junyou Li, Zhongbin Fang, Sheng Chen, Yangbin Yu, Qiang Fu, Wei Yang, and Deheng Ye.
\newblock Playable game generation.
\newblock \emph{arXiv preprint arXiv:2412.00887}, 2024{\natexlab{a}}.

\bibitem[Yang et~al.(2024{\natexlab{b}})Yang, Walker, Parker-Holder, Du, Bruce, Barreto, Abbeel, and Schuurmans]{pmlr-v235-yang24z}
Sherry Yang, Jacob~C Walker, Jack Parker-Holder, Yilun Du, Jake Bruce, Andre Barreto, Pieter Abbeel, and Dale Schuurmans.
\newblock Position: Video as the new language for real-world decision making.
\newblock In Ruslan Salakhutdinov, Zico Kolter, Katherine Heller, Adrian Weller, Nuria Oliver, Jonathan Scarlett, and Felix Berkenkamp, editors, \emph{Proceedings of the 41st International Conference on Machine Learning}, volume 235 of \emph{Proceedings of Machine Learning Research}, pages 56465--56484. PMLR, 21--27 Jul 2024{\natexlab{b}}.

\bibitem[Yang et~al.(2024{\natexlab{c}})Yang, Teng, Zheng, Ding, Huang, Xu, Yang, Hong, Zhang, Feng, et~al.]{yang2024cogvideox}
Zhuoyi Yang, Jiayan Teng, Wendi Zheng, Ming Ding, Shiyu Huang, Jiazheng Xu, Yuanming Yang, Wenyi Hong, Xiaohan Zhang, Guanyu Feng, et~al.
\newblock Cogvideox: Text-to-video diffusion models with an expert transformer.
\newblock \emph{arXiv preprint arXiv:2408.06072}, 2024{\natexlab{c}}.

\bibitem[Yu et~al.(2025)Yu, Qin, Wang, Wan, Zhang, and Liu]{yu2025gamefactory}
Jiwen Yu, Yiran Qin, Xintao Wang, Pengfei Wan, Di~Zhang, and Xihui Liu.
\newblock Gamefactory: Creating new games with generative interactive videos.
\newblock \emph{arXiv preprint arXiv:2501.08325}, 2025.

\bibitem[Yu et~al.(2023)Yu, Cheng, Sohn, Lezama, Zhang, Chang, Hauptmann, Yang, Hao, Essa, et~al.]{yu2023magvit}
Lijun Yu, Yong Cheng, Kihyuk Sohn, Jos{\'e} Lezama, Han Zhang, Huiwen Chang, Alexander~G Hauptmann, Ming-Hsuan Yang, Yuan Hao, Irfan Essa, et~al.
\newblock Magvit: Masked generative video transformer.
\newblock In \emph{Proceedings of the IEEE/CVF Conference on Computer Vision and Pattern Recognition}, pages 10459--10469, 2023.

\bibitem[Zhang et~al.(2018)Zhang, Isola, Efros, Shechtman, and Wang]{zhang2018lpips}
Richard Zhang, Phillip Isola, Alexei~A Efros, Eli Shechtman, and Oliver Wang.
\newblock The unreasonable effectiveness of deep features as a perceptual metric.
\newblock In \emph{Proceedings of the IEEE conference on computer vision and pattern recognition}, pages 586--595, 2018.

\bibitem[Zhang et~al.(2023)Zhang, Wang, Sun, Yuan, and Huang]{zhang2023storm}
Weipu Zhang, Gang Wang, Jian Sun, Yetian Yuan, and Gao Huang.
\newblock {STORM}: Efficient stochastic transformer based world models for reinforcement learning.
\newblock In \emph{Thirty-seventh Conference on Neural Information Processing Systems}, 2023.
\newblock URL \url{https://openreview.net/forum?id=WxnrX42rnS}.

\bibitem[Zhang et~al.(2024)Zhang, Bai, Zhao, Yan, Li, and Li]{zhang2024marie}
Yang Zhang, Chenjia Bai, Bin Zhao, Junchi Yan, Xiu Li, and Xuelong Li.
\newblock Decentralized transformers with centralized aggregation are sample-efficient multi-agent world models.
\newblock \emph{arXiv preprint arXiv:2406.15836}, 2024.

\bibitem[Zheng et~al.(2024)Zheng, Peng, Yang, Shen, Li, Liu, Zhou, Li, and You]{opensora}
Zangwei Zheng, Xiangyu Peng, Tianji Yang, Chenhui Shen, Shenggui Li, Hongxin Liu, Yukun Zhou, Tianyi Li, and Yang You.
\newblock Open-sora: Democratizing efficient video production for all, March 2024.
\newblock URL \url{https://github.com/hpcaitech/Open-Sora}.

\bibitem[Zhu et~al.(2024)Zhu, Wu, Guo, Liu, Cheang, and Kong]{FangqiIRASim2024}
Fangqi Zhu, Hongtao Wu, Song Guo, Yuxiao Liu, Chilam Cheang, and Tao Kong.
\newblock Irasim: Learning interactive real-robot action simulators.
\newblock \emph{arXiv:2406.12802}, 2024.

\end{thebibliography}
\bibliographystyle{plainnat}

\newpage
\appendix
\section{Details of Dataset Collection}
\label{appendix:dataset}
\subsection{Atari}
In our experiments on two Atari games - \emph{Breakout} and \emph{Battle Zone}, we utilized a well-established baseline for Atari, Deep Q-Learning (DQN) agent \citep{mnih2013dqn}, to collect the offline dataset for tuning. Specifically, we adopted an open-source DQN implementation provided at \url{https://github.com/vwxyzjn/cleanrl}, trained the agent for 1 million environment steps, and stored the replay buffer as the offline dataset. Since our work focuses on developing a more accurate world model, we intentionally initialized these two game environments without enabling random sticky actions. This ensures that the collected trajectories accurately capture the ground-truth state transitions of the environment. Further, we processed the initialized environments by applying a FrameSkipping wrapper with a frame skip of 4 and a ResizingObservation wrapper, which resizes the output RGB observations to a resolution of $(84, 84)$ for DQN agent training. In terms of the hyperparameter of DQN, we followed the default setting provided at \url{https://github.com/vwxyzjn/cleanrl/blob/master/cleanrl/dqn_atari.py}.
\subsection{Procgen}
Following a similar approach to our Atari collection experiments, we collected replay datasets from two Procgen games: Coinrun and Ninja. The data collection process utilized a publicly available Proximal Policy Optimization (PPO) implementation~\citep{shengyi2022the37implementation}, accessed through \url{https://github.com/vwxyzjn/cleanrl/blob/master/cleanrl/ppo_procgen.py}. For each game, we collected 1 million state transitions at the default Procgen resolution of $64\times64$ pixels for DWS training.
\section{Implementation Details}
\label{appendix:details}
\subsection{Details of Base Models}
\label{appendix:base_models}
\paragraph{Open-Sora.} Open-Sora, based on rectified-flow~\citep{liu2022flow} and spatial-temporal transformer~\citep{ma2024latte,chen2023pixartalpha} architecture , is a text-and-frame-conditioned video generation model. For our implementation of DWS, we utilized Open-Sora version 1.2 as our base model. To maintain consistency with the original architecture, we employed the same Variational Autoencoder (VAE) provided in the Open-Sora library. However, to optimize computational efficiency, we adopted T5-small for text encoding instead of the original text encoder. Given that the original Open-Sora model comprises 1.1 billion parameters, which poses significant computational constraints for inference and downstream applications, we strategically initialized our model using only the first 12 layers of the Open-Sora architecture. This modification substantially reduced the model's computational footprint while preserving essential generative capabilities. Our proposed action-conditioned module is integrated into each layer of Open-Sora to improve action-frame alignment.

\paragraph{iVideoGPT.} iVideoGPT is an autoregressive transformer-based architecture that builds upon the LLaMA architecture~\citep{touvron2023llama}. The model extends the base architecture by incorporating specialized reward prediction head-layers, enabling it to perform both video generation and reward estimation tasks. A distinctive feature of iVideoGPT is its context-aware tokenizer, which implements compressive tokenization for efficient video prediction. We initialize our model with weights from \url{https://huggingface.co/thuml/ivideogpt-oxe-64-act-free}, including a tokenizer of 114M size and a transformer of 138M size. Our proposed action-conditioned module is integrated into each transformer block.

\subsection{Implementation of Motion-Reinforced Loss}
\label{appendix:motion_loss}
For diffusion-based models, which use a squared $\mathit{l}_2$ distance for computing loss function and conducting gradient updates, their original loss functions can be represented as follows:
\begin{equation}
    \mathcal{L}_{\rm prev} = \mathbb{E}[\Vert g-p_{\theta}(x_t,t,y)\Vert^2_2],
\end{equation}
where $y$ is the condition, and $g$ represents the target signal to be estimated, which takes different forms depending on the model architecture: For DDPM-style diffusion models~\citep{ddpm}, it corresponds to the ground-truth noise, while for flow-matching-based models~\citep{liu2022flow}, it represents the velocity. Given that the output of $p_{\theta}$ and the video embedding $x_t$ share identical dimensionality, thus we can incorporate the motion-based weights $\omega$ into $\mathcal{L}_{\rm prev}$ as follows:
\begin{equation}
    \mathcal{L}_{\rm motion}=\mathbb{E}[\omega\Vert g-p_{\theta}(x_t,t,y)\Vert^2_2].
\end{equation}
For autoregressive transformer-based models, which use a cross-entropy loss for training, their original loss functions can be represented as follows:
\begin{equation}
    \mathcal{L}_{\rm prev} =-\sum_{i=t}^{L}\log p(x_i|x_{<i}),
\end{equation}
where $x_i$ denotes discrete tokens embedded in the transformer, and $L$ is the sequence length. In this case, we compute $\omega$ as follows:
\begin{equation}
    \omega_{i+1} = c^{\mathbb{I}(x_{i+1}=x_{i})},
\end{equation}
where $c=e$ and $\mathbb{I}$ denotes an indicator function. Thus, we obtain the final $\mathcal{L}_{\rm motion}$ for transformer-based models:
\begin{equation}
    \mathcal{L}_{\rm motion}=-\sum_{i=t}^{L}\omega \log p(x_i|x_{<i}),
\end{equation}
\subsection{Implementation of the MBRL Algorithm}
We develop a simple model-based RL algorithm using a DWS-trained world model within the MBPO framework~\citep{janner2019mbpo,wu2024ivideogpt}, with PPO~\citep{schulman2017proximal} as the base actor-critic RL algorithm. We provide the pseudo-code in Alg.~\ref{alg:mbpo}. We build our codes upon the implementation in \url{https://github.com/vwxyzjn/cleanrl/blob/master/cleanrl/ppo_procgen.py} for Procgen environments and \url{https://github.com/vwxyzjn/cleanrl/blob/master/cleanrl/ppo_atari.py} for Atari environments, using the same hyperparameters and architecture for actor-critic learning.
Hyperparameters specific to model-based RL are listed in Table~\ref{tab:hyper_mbrl}. Following Dreamerv3~\citep{hafner2023dreamerv3}, we use a symlog transformation for reward prediction.
\label{appendix:mbrl}
\begin{algorithm}[htbp]
  \caption{Model-Based Reinforcement Learning~\citep{janner2019mbpo} with Prioritized Imagination}
  \label{alg:mbpo}
  \begin{algorithmic}[1]
    \STATE Initialize real replay buffer $\mathcal{B}_\text{real}$ with random policy
    \STATE Initialize actor-critic $\pi_{\phi}, v_{\phi}$, world model $p_{\theta}$
    \STATE Initially train model $p_{\theta}$ on $\mathcal{B}_\text{real}$
    \STATE Initialize imagined replay buffer $\mathcal{B}_\text{imag}$
    \FOR{$N$ steps}
      \STATE {\texttt{// \textcolor{gray}{Training}}}
      \IF {world model update}
        \STATE Update world model $p_{\theta}$ on a mini-batch from $\mathcal{B}_\text{real}$
      \ENDIF
      \STATE Compute TD-error $\delta$, loss $\mathcal{L}_{\rm rl}$ with PPO on a mini-batch from $\mathcal{B}_\text{imag}\cup \mathcal{B}_\text{real}$
      \STATE Update actor-critic $\pi_{\phi}, v_{\psi}$
      \STATE Update priorities of samples in mini-batch from $\mathcal{B}_\text{real}$ with the new values as $max(\delta,0)+\epsilon$ \\ \texttt{// \textcolor{gray}{$\epsilon=1e^{-6}$}}
      \STATE {\texttt{// \textcolor{gray}{Data collection}}}
      \IF {world model rollout}
          \STATE Sample a mini-batch of $o_t$ from $\mathcal{B}_\text{real}$ based on priorities
          \STATE Perform $k$-step model rollout starting from $o_t$ using policy $\pi_{\phi}$; add to $\mathcal{B}_\text{imag}$
      \ENDIF
        \STATE Take action in environment according to $\pi_{\phi}$; add to $\mathcal{B}_\text{real}$
    \ENDFOR
  \end{algorithmic}
\end{algorithm}
\begin{table}[htbp]
    \centering
    \begin{tabular}{c|c}
         Hyperparameter& Value \\
         World model rollout batch size& 16\\
         World model rollout horizon & 9\\
         Real data ratio & 0.5\\
         World model training batch size&16\\
         World model training learning rate&2e-5\\
         Optimizer& Adam
    \end{tabular}
    \caption{Hyperparameters of our proposed model-based RL algorithm with prioritized imagination.}
    \label{tab:hyper_mbrl}
\end{table}

\end{document}